\documentclass[10pt,twocolumn,letterpaper]{article}

\usepackage[pagenumbers]{iccv} 

\usepackage{algorithm}
\usepackage{algorithmic}
\usepackage{multirow}
\usepackage{colortbl}
\usepackage{cuted}
\definecolor{iccvblue}{rgb}{0.21,0.49,0.74}
\usepackage[pagebackref,breaklinks,colorlinks,allcolors=iccvblue]{hyperref}

\def\paperID{6175} 
\def\confName{ICCV}
\def\confYear{2025}

\title{Leveraging Local Patch Alignment to Seam-cutting for Large Parallax Image Stitching}

\author{Tianli Liao, Chenyang Zhao, Lei Li, Heling Cao\\
College of Information Science and Engineering, Henan University of Technology, China\\
{\tt\small \{tianli.liao,leili,caohl\}@haut.edu.cn, zhaochy2005@163.com}
}

\begin{document}

\maketitle


\begin{abstract}
Seam cutting has shown significant effectiveness in the composition phase of image stitching, particularly for scenarios involving parallax. However, conventional implementations typically position seam-cutting as a downstream process contingent upon successful image alignment. This approach inherently assumes the existence of locally aligned regions where visually plausible seams can be established. Current alignment methods frequently fail to satisfy this prerequisite in large parallax scenarios despite considerable research efforts dedicated to improving alignment accuracy. In this paper, we propose an alignment-compensation paradigm that dissociates seam quality from initial alignment accuracy by integrating a Local Patch Alignment Module (LPAM) into the seam-cutting pipeline. Concretely, given the aligned images with an estimated initial seam, our method first identifies low-quality pixels along the seam through a seam quality assessment, then performs localized SIFT-flow alignment on the critical patches enclosing these pixels. Finally, we recomposite the aligned patches using adaptive seam-cutting and merge them into the original aligned images to generate the final mosaic. Comprehensive experiments on large parallax stitching datasets demonstrate that LPAM significantly enhances stitching quality while maintaining computational efficiency. The code is available at \url{https://github.com/tlliao/LPAM_seam-cutting}.
\end{abstract}

\section{Introduction}

Image stitching, a cornerstone technology in computer vision with decades of development \cite{szeliski2006image}, fundamentally comprises two critical phases: image alignment \cite{Brown:2007,gao2011constructing,zaragoza2014projective,li2018parallax,herrmann2018robust,li2019local,Zheng2019tmm,Lee_2020_CVPR,li2020local,zhang2021natural,zhang2022Image,chen2024seamless}, composition and image blending \cite{kwatra2003graphcut,zhang2016multi,herrmann2018object,li2018perception,liao2019Quality,zhang2025seam}. Despite remarkable progress, parallax artifacts persist as the Achilles heel in consumer-level photography due to unconstrained capture conditions. Current alignment frameworks frequently fail to address this challenge adequately, making the composition stage ultimately decisive for output quality.

\begin{figure*}
	\centering
	\subfloat[\centering Input images]{
		\includegraphics[width=0.303\textwidth]{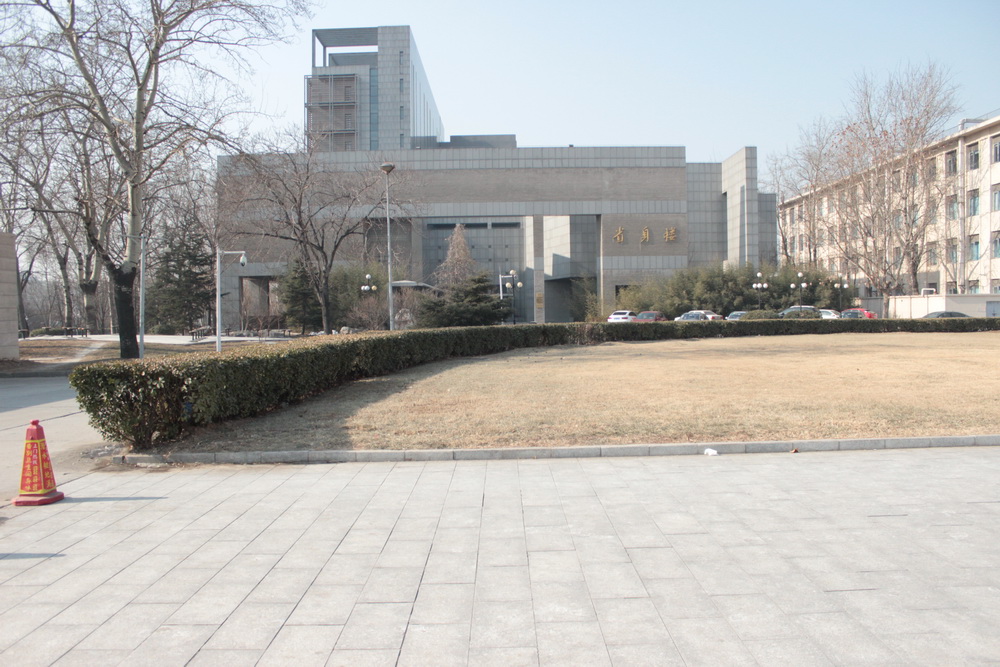}
		\includegraphics[width=0.303\textwidth]{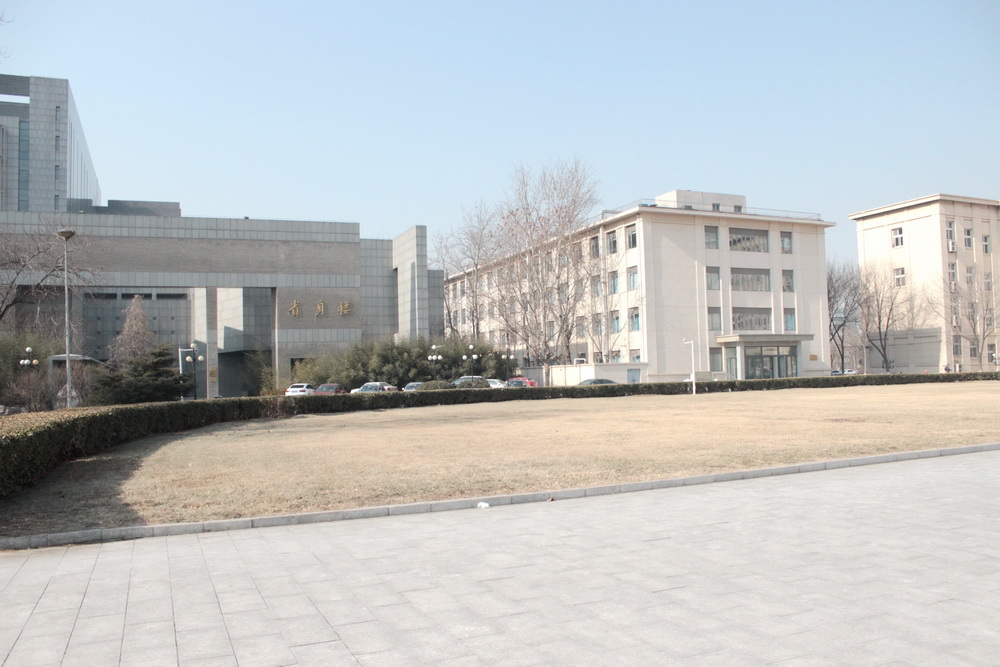}}
	\subfloat[\centering Averaged aligned images]{
		\includegraphics[width=0.354\textwidth]{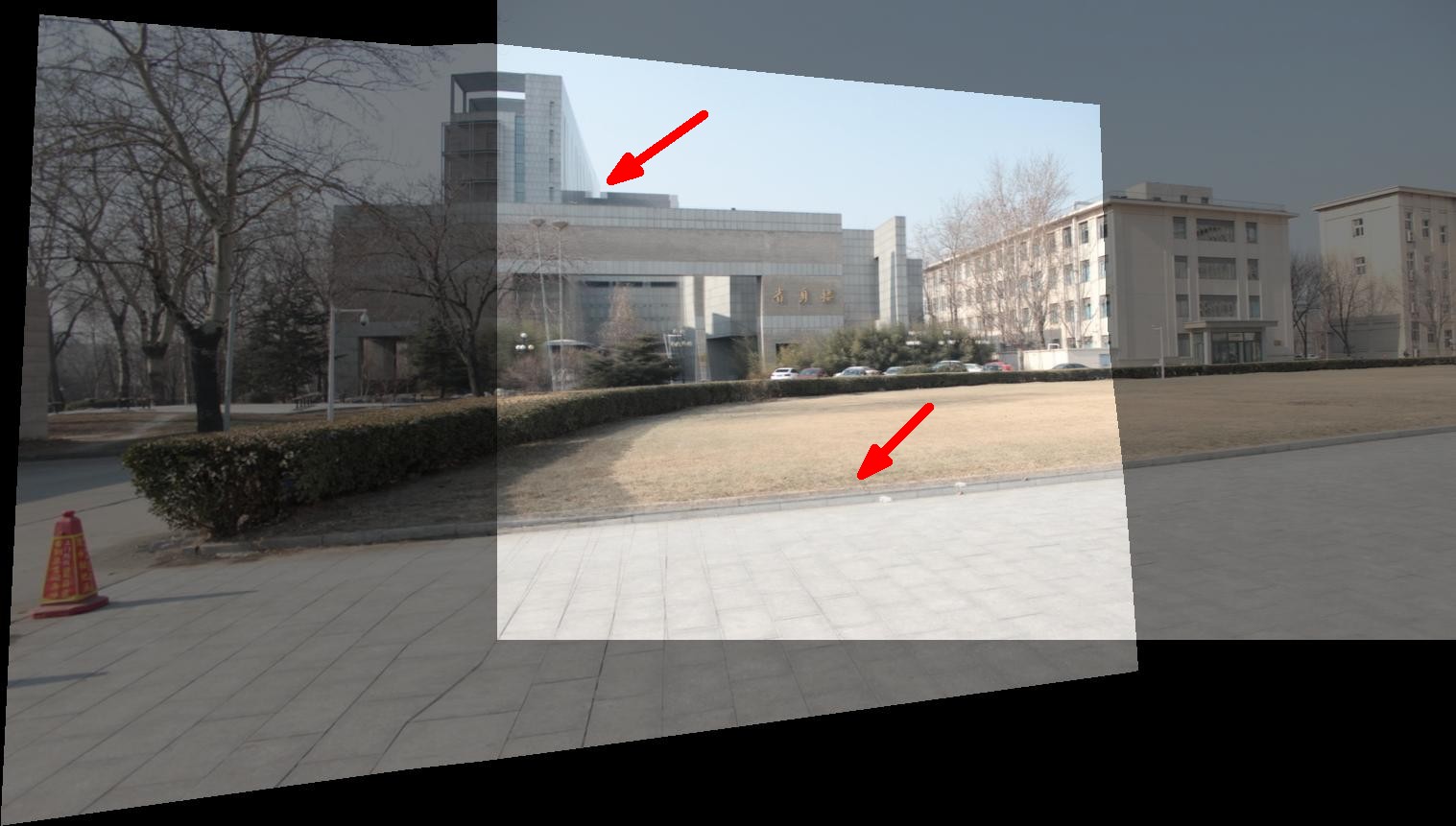}}\\
	\subfloat[\centering Method \cite{liao2019Quality}]{
		\includegraphics[width=0.32\textwidth]{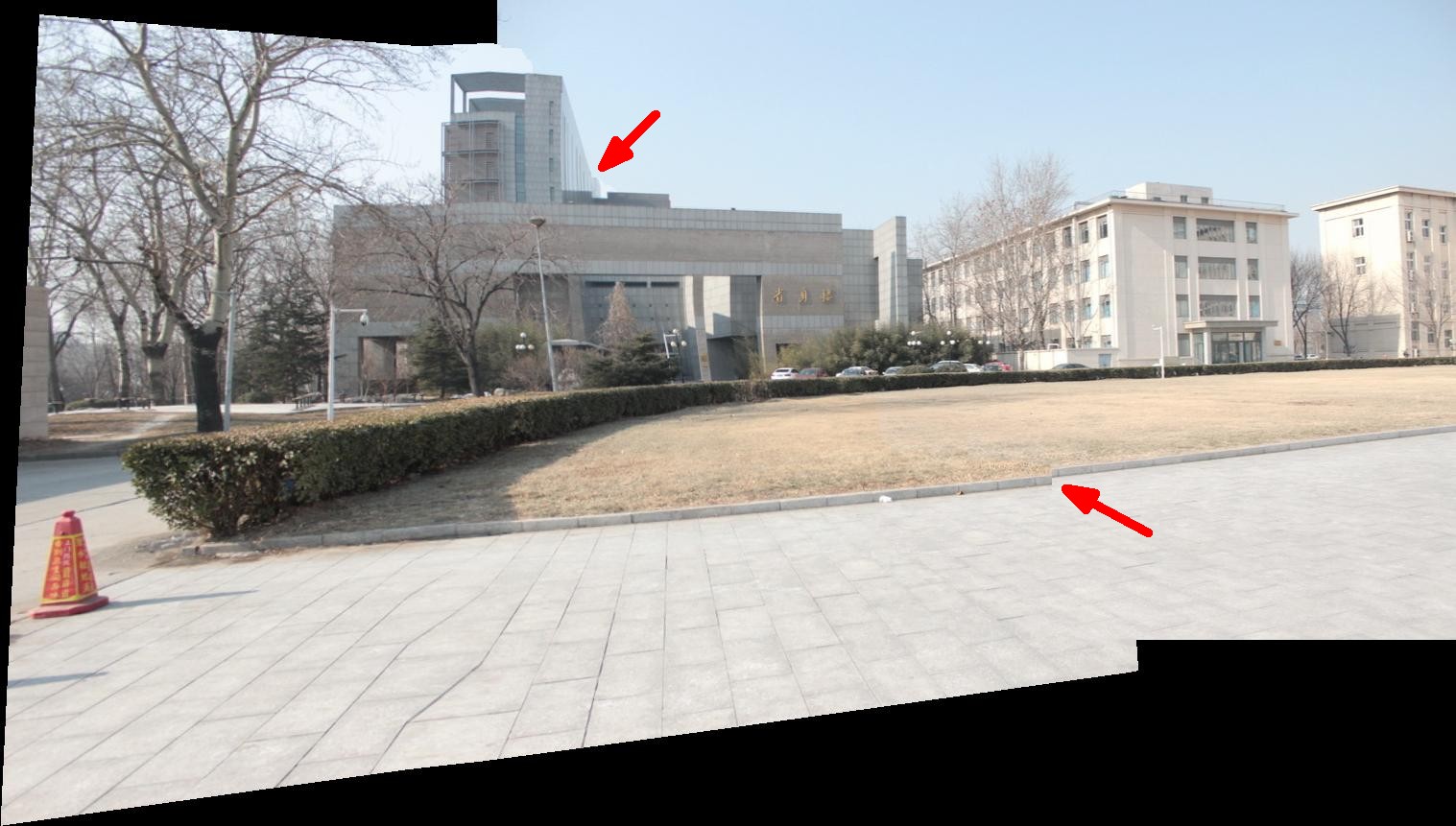}}			
	\subfloat[\centering Method \cite{li2024automatic}]{
		\includegraphics[width=0.32\textwidth]{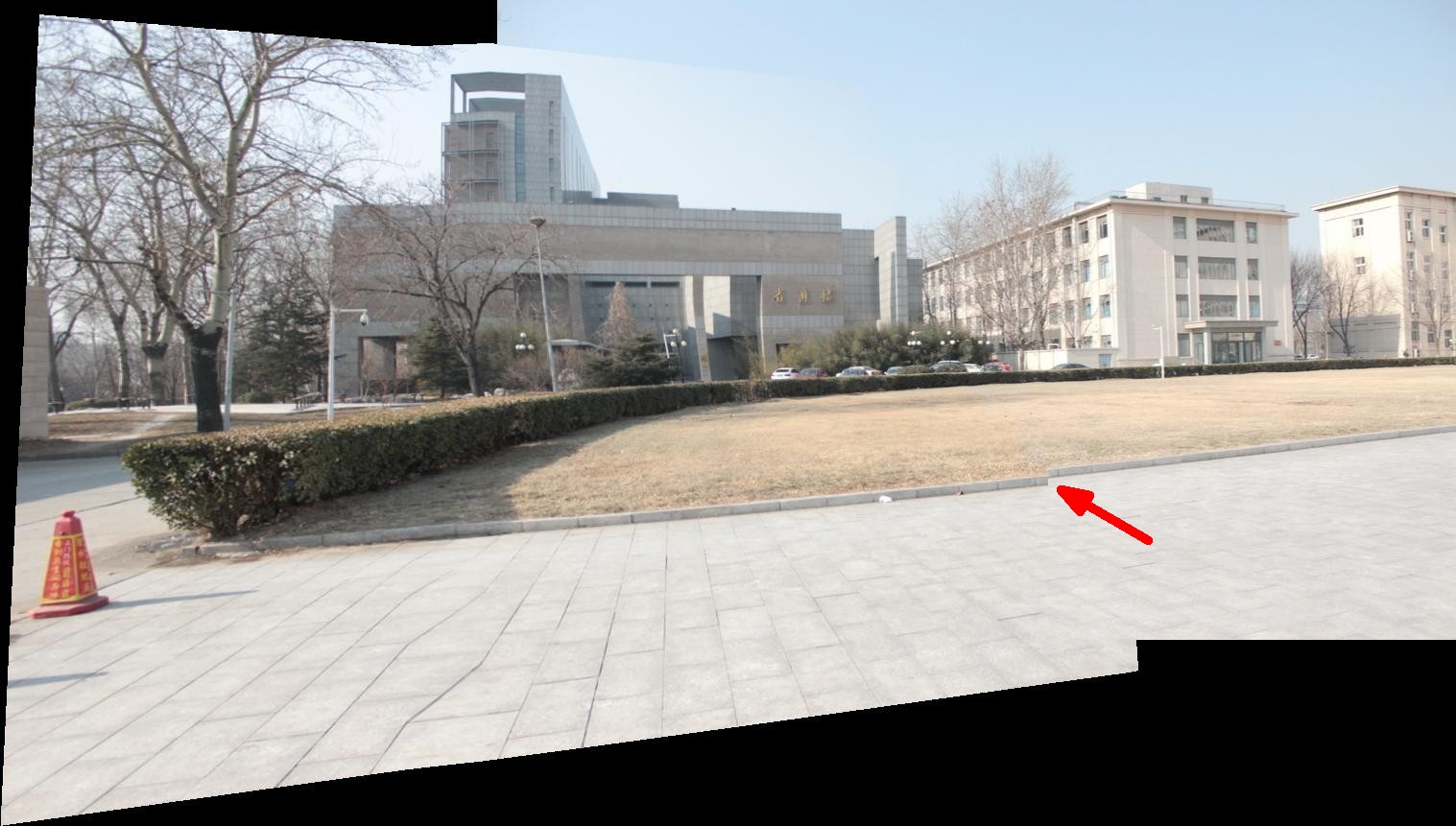}}
	\subfloat[\centering Method \cite{liao2019Quality} + Our LPAM]{
		\includegraphics[width=0.32\textwidth]{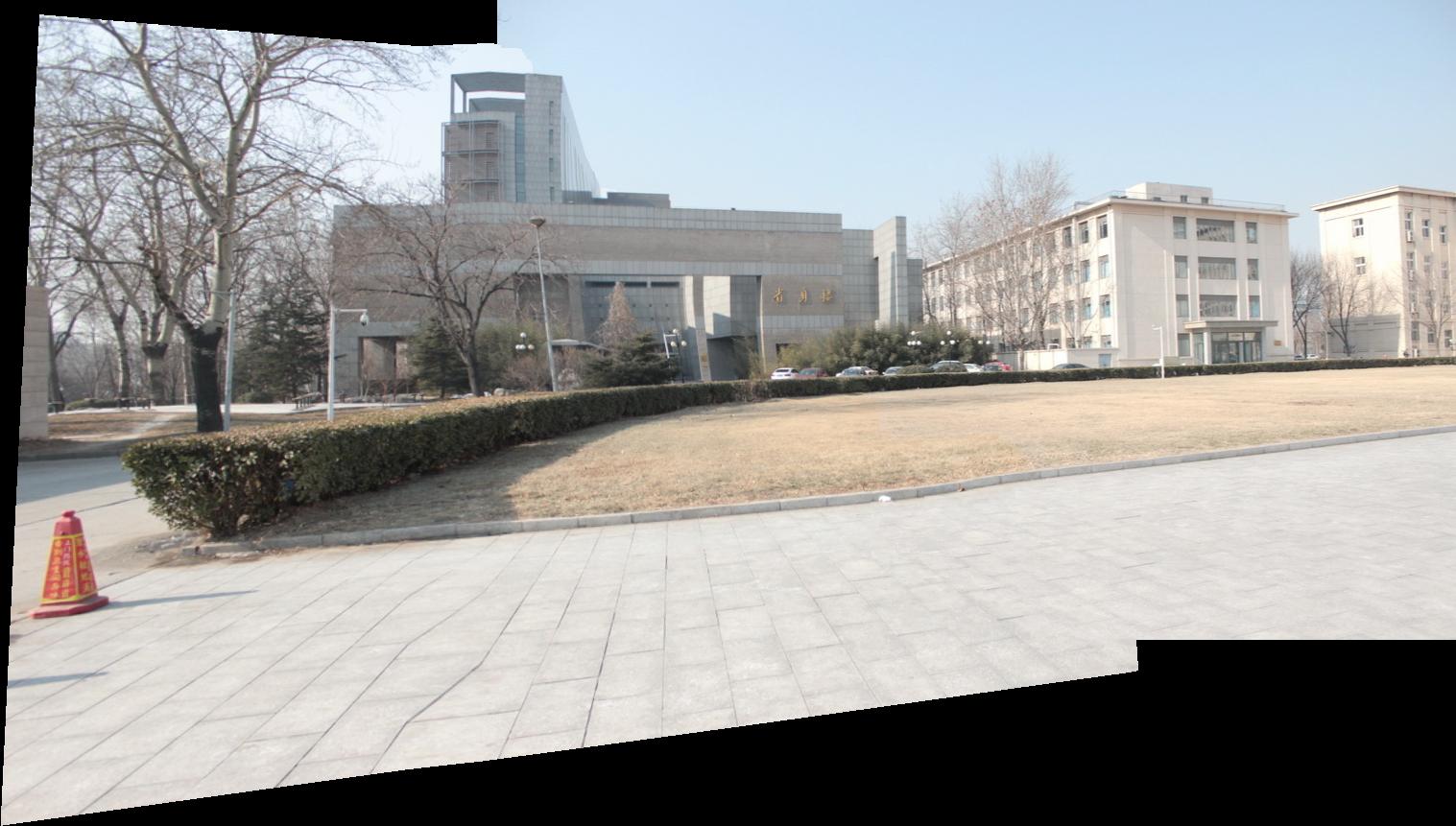}}
	\caption{Issues of the state-of-the-art seam-cutting methods for large parallax image stitching (marked by red arrows). (b) Aligned images are obtained via method \cite{li2018parallax}, where the overlapping region is averaged. The curbs are notably misaligned (Best view in color and zoom in).}
	\label{fig:1}
\end{figure*}

Contemporary solutions predominantly employ two strategies: \textit{seam-cutting} \cite{li2018perception,herrmann2018object,liao2019Quality,dai2021edge,li2024automatic,zhang2025seam} identifying optimal transition boundaries, and \textit{seam-driven stitching} \cite{gao2013seam,zhang2014parallax,lin2016seagull} choosing an optimal seam from multiple alignment candidates. However, these post-alignment approaches inherit fundamental limitations: their effectiveness strictly depends on initial alignment accuracy. As shown in Fig. \ref{fig:1}(b), significant parallax causes local misalignment that conventional seam-cutting methods cannot rectify, leading to visually disruptive artifacts (Fig. \ref{fig:1}(c,d)).

Our key insight is that localized \textit{ post hocc} alignment augmentation can overcome this bottleneck. We propose the Local Patch Alignment Module (LPAM), a lightweight add-on for existing seam-cutting pipelines. Specifically, given aligned images with an initial seam, we first identify low-quality pixels along the seam through a seam quality metric. Then, we separate their enclosing patches in the aligned images and perform localized SIFT-flow alignment for the patches. Finally, we recomposite the aligned patches using adaptive seam-cutting and merge them into the original aligned images to generate the final mosaic. We conduct extensive experiments to validate that adding LPAM into a seam-cutting can effectively and efficiently improve the stitching results, as shown in Fig. \ref{fig:1}(e).
Our core contributions are threefold:
\begin{itemize}
	\item A novel \textit{alignment-compensation} paradigm that dissociates seam quality from initial alignment accuracy, overcoming a fundamental limitation in current methodologies
	\item The first modular framework enabling seamless integration of local alignment refinement into arbitrary seam-cutting pipelines.
	\item An effective implementation achieving stitching quality improvement without requiring architectural changes to host algorithms.
\end{itemize}

\section{Related Work}

We briefly review the approaches most related to our work. Comprehensive reviews of image stitching algorithms are given in \cite{szeliski2006image,adel2014image,yan2023deep}.

\subsection{Seam-cutting methods}

The seam-cutting approach is a powerful and widely used composition method \cite{kwatra2003graphcut,agarwala2004interactive,zhang2016multi,herrmann2018object,li2018perception,liao2019Quality,miao2023superpixel}, which is usually expressed in terms of energy minimization and minimized via graph-cut optimization \cite{boykov2001fast}. Agarwala \emph{et al.} \cite{agarwala2004interactive} applied seam-cutting to generate a ``digital photomontage''. They defined the energy function as the sum of a data penalty selected by the user and an interaction penalty involving the colors, gradients, and edges of an image. 

Zhang \emph{et al.} \cite{zhang2016multi} combined the alignment error and color difference in the energy function to find a better seam that avoids splitting misaligned pixels with similar colors. Li and Liao \cite{li2018perception} considered the non-linearity and non-uniformity of human perception into seam-cutting and defined a sigmoid-metric color difference and saliency weights in the energy function. To avoid objects from being cropped, omitted, or duplicated, Herrmann \emph{et al.} \cite{herrmann2018object} used object detection to match objects between images and introduced three new terms to the traditional energy function that address the above issues, respectively. Liao \emph{et al.} \cite{liao2019Quality} indicated that the seam with minimal cost is not necessarily optimal in terms of human perception. Then, they proposed an iterative seam estimation method where a quality evaluation algorithm for pixels along the seam is designed to guide the iteration procedure. Dai \emph{et al.} \cite{dai2021edge} proposed the first end-to-end deep learning framework for image composition. They cast the seam-cutting as an image-blending problem and designed an edge-guided network to regress the blending weights and seamlessly produce the stitched image. Li and Zhou \cite{li2024automatic} utilize the quaternion representation of the color image to define the energy function and perform seam-cutting completely in the quaternion domain. 

\subsection{Seam-driven methods}

\begin{figure*}[t]
	\centering
	\includegraphics[width=0.95\textwidth]{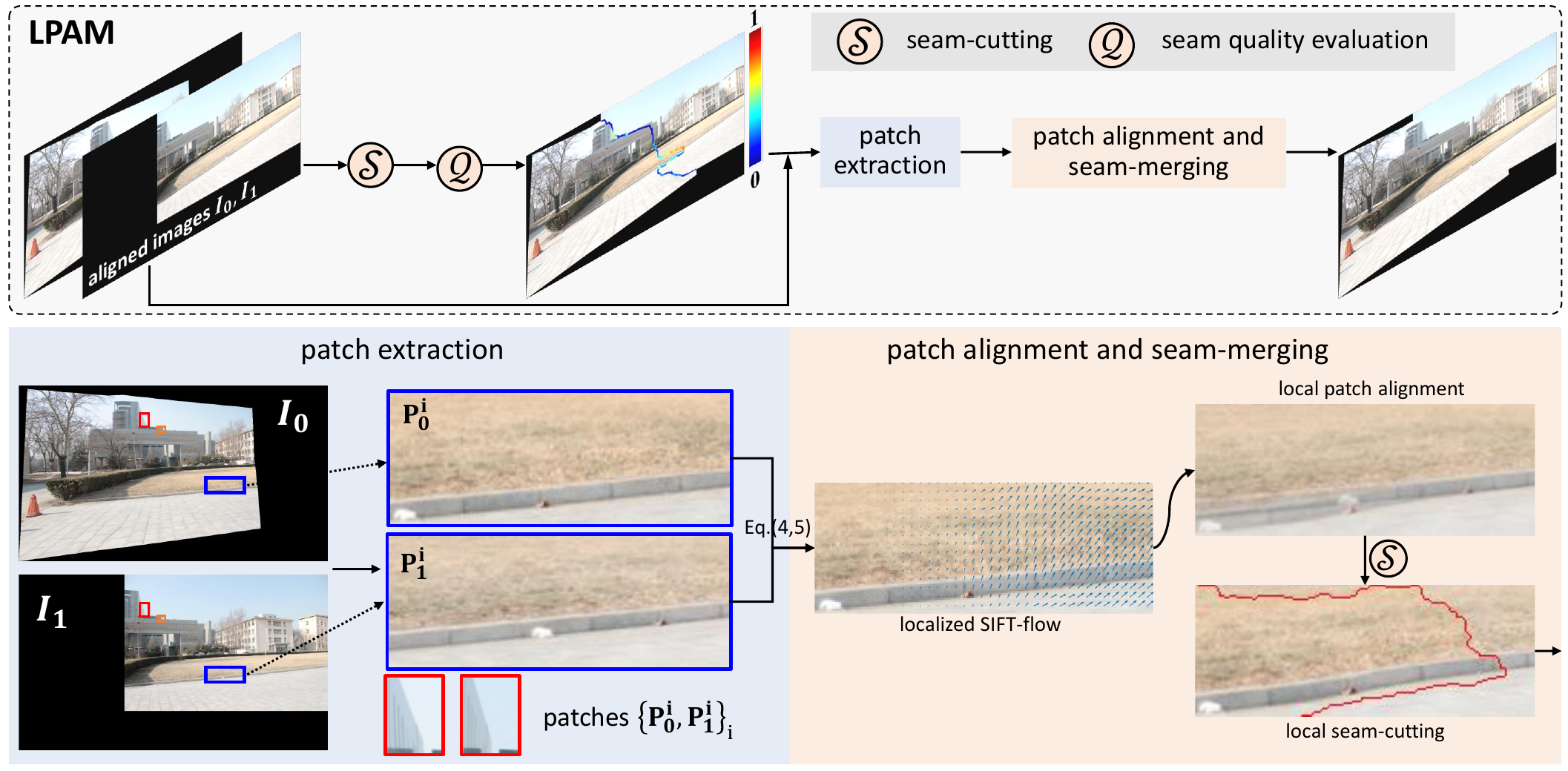}\\
	\caption{An overview of the proposed local patch alignment module. Given two aligned images, we first estimate an initial seam and evaluate the pixels' quality along the seam. Then, we adopt two stages to address the low-quality pixels: (1) Extract their enclosing patches in the aligned images for further realignment. (2) Realign each two patches using localized SIFT-flow, composite the aligned patches via a local seam-cutting.}
	\label{fig: pipeline}
\end{figure*}

The seam-driven strategy was first proposed by Gao \emph{et al.} \cite{gao2013seam}. They indicated that the perceptually best result is not necessarily from the best global alignment. Thus, they generated finite alignment candidates from multiple RANSAC \cite{fischler1981random} procedures and applied seam-cutting to estimate multiple seams from these candidates. Then, a seam quality metric is defined to evaluate these seams, and the seam with minimal cost is chosen to produce a final mosaic. Zhang and Liu \cite{zhang2014parallax} improved the screening process of alignment candidates by combining homography with content-preserving warps. They used the seam cost of the energy function to evaluate the seam quality. Lin \emph{et al.} \cite{lin2016seagull} generated the alignment candidates via a superpixel-based feature grouping and a structure-preserving warp. The warp is gradually improved via adaptive feature weighting. They also defined a seam quality metric based on the zero-mean normalized cross-correlation (ZNCC) score. Zhang \emph{et al.} \cite{zhang2025seam} proposed an iterative seam-based stitching method that leverages dense flow estimation to directly estimate the seam in the input images and use it to solve a spatial smooth warp.

Both seam-cutting and seam-driven methods focus on defining various energy functions to find an optimal seam from the settled pre-aligned images. For images with large parallax, these methods may easily fail if the alignment methods fail to align the images locally. Our method, on the other hand, improves performance by realigning the locally misaligned patches where the estimated seam passes.

\section{Proposed Method}

As depicted in Fig. \ref{fig: pipeline}, our method consists of two main stages, seam quality evaluation and patch extraction, succeeded by patch alignment and seam-merging. Subsequently, we first provide a preliminary of the basic seam-cutting algorithm and then present a detailed description of the two stages.

\subsection{Preliminary}

Given aligned target image $I_0$ and reference image $I_1$, for the overlapping region $\mathbf{O}$, seam-cutting method estimates an initial seam $S_0$ by finding a labeling $l$ that minimizes the energy function
\begin{equation}\label{eq:seam-cutting}
	E(l) = \sum_{p\in\mathbf{O}}D_p(l_p)+\sum_{(p,q)\in N}S_{p,q}(l_p,l_q)
\end{equation}
where the data penalty $D_p$ measures the cost of assigning a label $l_p$ to pixel $p$, the smoothness penalty $S_{p,q}$ measures the cost of assigning a pair of labels $(l_p, l_q)$ to a pair of neighboring pixels $(p, q)\in N$. Eq. (\ref{eq:seam-cutting}) can be minimized via graph-cut optimization \cite{boykov2001fast}.

\subsection{Seam quality evaluation and patch extraction}

Since the seam-cutting methods mainly fail due to the structure misalignment, we use the structural similarity (SSIM) index \cite{wang2004image} to evaluate each pixel $p_i$ on the seam $S_0$. Concretely, we extract two local patches $P_0$ and $P_1$ centered at $p_i$, and define the evaluation error for $p_i$ as
\begin{equation}\label{eq:ssim}
	Q(p_i) = 1-\mathrm{SSIM}(P_0(p_i), P_1(p_i)),
\end{equation}
Fig. \ref{fig: pipeline} illustrates one example of the seam quality evaluation for seam $S_0$, pixels with high errors usually mean misaligned parts.

\begin{figure*}[t]
	\centering
	\includegraphics[width=0.98\textwidth]{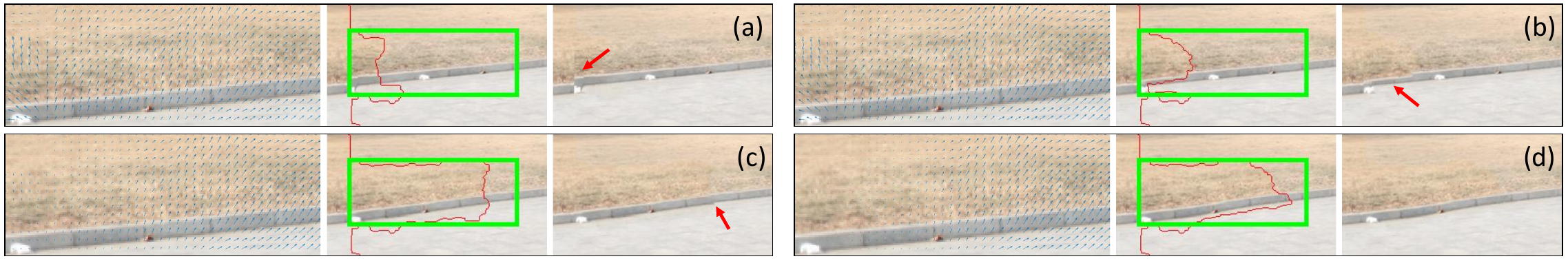}\\
	\caption{Illustration for patch alignment and seam-merging process. Artifacts are denoted by red arrows. From left to right in (a,b,c,d): vector flow, local stitching seam, and local composited result. For clarity, only part of the vector flow is presented. From (a) to (d): results of original $V^i$, modified $f(V^i)$ with $\beta=1,4,8$ in Eq. (\ref{eq:sigmoid}), respectively. The green rectangle in the middle of each subfigure indicates the position of the aligned patch and the merging position of the local seam. The right image in each subfigure shows the composited patch result via the seam-cutting in the middle (Best view in color and zoom in).}
	\label{fig:siftflow}
\end{figure*}

Due to the advanced alignment methods, even for images with large parallax, the seam only experiences partial failure.  Our goal is to ``repair'' these failed parts. As noted in \cite{liao2019Quality}, a plausible seam should have a relatively small error $Q$ for every pixel. Thus, for pixels with high errors, we separate them in the seam for subsequent alignment as follows,
\begin{itemize}
	\item if $\max(Q)\leq k \cdot \mathrm{mean}(Q)$, we consider $S_0$ as a plausible seam;
	\item else, we calculate a threshold $\tau$ via Ostu's method \cite{otsu1975threshold}, and consider $p_i$ with $Q(p_i)\geq\tau$ as misaligned.
\end{itemize}
We set $k$ to 1.5 in our method. The threshold $\tau$ divides the seam into multiple components. Notably, the threshold might not be sufficiently precise to identify all misaligned pixels. However, we allow for some false positives because the subsequent alignment process does not yield inferior results. For each misaligned component $\mathbf{C}_i$, we separate its enclosing rectangular patches $\mathbf{P}^i_0$ and $\mathbf{P}^i_1$ from the corresponding aligned images (see Fig. \ref{fig: pipeline}).

\subsection{Patch alignment and seam-merging}

\subsubsection{Patch alignment}
The rectangular patches manifest structural misalignment stemming from the feature-based alignment algorithm, primarily due to insufficient feature correspondences in texture-deficient regions.
Thus, we utilize SIFT flow \cite{Liu2011SIFTflow} to construct dense correspondences among the pixels within the patches. Theoretically, for each pixel $p_j$ in $\mathbf{P}_0^i$, there exists a shift vector $V^i(p_j)$ such that
\begin{equation}\label{eq:siftflow}
	\|\mathbf{P}_0^i(p_j + V^i(p_j))-\mathbf{P}_1^i(p_j)\|=0
\end{equation}
Then, the two patches can be aligned via the vector flow $V^i$. However, if the boundary pixels of the patch $\mathbf{P}_0^i$ have large shift vectors, merging the aligned patch into the initial aligned result may introduce more artifacts around the patch, as shown in Fig. \ref{fig:siftflow}. To address this, we smoothly modify the vector flow and realign the two patches as 
\begin{equation}\label{eq:siftflow_1}
	\tilde{\mathbf{P}}_0^i(p_j)=\mathbf{P}_0^i(p_j + f(p_j)\cdot V^i(p_j)),
	\tilde{\mathbf{P}}_1^i(p_j)=\mathbf{P}_1^i(p_j)
\end{equation}
where $\tilde{\mathbf{P}}_0^i$ ($\tilde{\mathbf{P}}_1^i$) is the warped target (reference) patch, $f$ is the sigmoid function
\begin{equation}\label{eq:sigmoid}
	f(t)=\frac{1}{1+e^{-\beta(t-0.5)}}
\end{equation}
For horizontal patch alignment, $t$ represents the $x$-coordinate of the pixel $p_j$ within patch $\mathbf{P}_0^i$, normalized to $[0,1]$. $\beta$ controls the rate of change of the function $f$. The sigmoid function ensures that the warped patch $\tilde{\mathbf{P}}_0^i$ is smoothly consistent with $I_0$ when $t=0$ and with $I_1$ when $t=1$. The warped
patches $\tilde{\mathbf{P}}_0^i$ and $\tilde{\mathbf{P}}_1^i$
are then placed back to the original aligned images $I_0$ and $I_1$. Fig. \ref{fig:siftflow} shows a comparison example of the modified vector flow under different settings of $\beta$. In our method, $\beta$ is set to 8.

\subsubsection{Seam-merging}

Due to the vector flow modification, the rectangular patches are only partially aligned. Thus, seam-cutting is still required to composite these patches. Note that the local seam $s_i$ in the aligned patch must be connected with the initial seam $S_0$ end to end. Otherwise, artifacts may occur at the junction of the two seams. To prevent introducing new artifacts, we constrain the local seam to connect with the initial seam by finding a label $l^i$ in which the data penalty for pixel $p_j\in\tilde{\mathbf{P}}_0^i$ is
\begin{itemize}
	\item $D_{p_j}(0)=0, D_{p_j}(1)=\infty$, if $p_j\in\partial \tilde{\mathbf{P}}_0^i$ with $l_{p_j}=0$ in the initial seam $S_0$;
	\item $D_{p_j}(0)=\infty, D_{p_j}(1)=0$, if $p_j\in\partial \tilde{\mathbf{P}}_0^i$ with $l_{p_j}=1$ in the initial seam $S_0$;
	\item $D_{p_j}(0)=0, D_{p_j}(1)=0$, otherwise,
\end{itemize}
where $\infty$ is set to avoid mislabeling, $\partial \tilde{\mathbf{P}}_0^i$ represents the border of the warped patch $\tilde{\mathbf{P}}_0^i$. Fig. \ref{fig:siftflow} shows an example including the seam-merging process. We can see that the local seam in the aligned patch is accurately merged with the initial seam.
After realigning all the rectangular patches, we can generate the final mosaic by compositing the locally realigned images $\tilde{I}_0$ and $\tilde{I}_1$ with the new stitching seam.
We summarize our method in Algorithm \ref{algor:1}.
\begin{algorithm}
	\caption{Local patch alignment module for large parallax image stitching}
	\label{algor:1}
	\begin{algorithmic}[1]
	\REQUIRE{aligned images $I^0_0$, $I^0_1$ and an initial seam $S_0$}
	\ENSURE{final seam $S_*$ and realigned images $I^*_0$, $I^*_1$}
	\STATE evaluate $S_0$ via Eq. (\ref{eq:ssim});
	\STATE identify whether there are misaligned pixels along $S_0$;
	\IF{misaligned pixels exist}
	\STATE extract misaligned components $\{\mathbf{C}_i\}_{i=1}^n$ in $S_0$;
		\FOR{$i=1$ \TO n}
			\STATE separate the enclosing patches $\mathbf{P}^i_0$ and $\mathbf{P}^i_1$ from $I_0$ and $I_1$, respectively;
			\STATE obtain $V_i$ via Eq. (\ref{eq:siftflow});
			\STATE modify the vector flow via $f$ in Eq. (\ref{eq:sigmoid});
			\STATE obtain $\tilde{\mathbf{P}}_0^i$ and $\tilde{\mathbf{P}}_1^i$ via Eq. (\ref{eq:siftflow_1});
			\STATE calculate a local stitching seam $s_i$ in $\tilde{\mathbf{P}}_0^i$ and $\tilde{\mathbf{P}}_1^i$;
			\STATE $S_i:=S_{i-1}\leftarrow s_i$;
			\STATE  $I^i_0:=I^{i-1}_0 \leftarrow \tilde{\mathbf{P}}_0^i$;
			\STATE $I^i_1:=I^{i-1}_1 \leftarrow \tilde{\mathbf{P}}_1^i$.
		\ENDFOR
	\ENDIF 
	\RETURN $S_*=S_n, I^*_0=I^n_0, I^*_1=I^n_1$ 
	\end{algorithmic}
\end{algorithm}


\begin{figure}[t]
	\centering
	\includegraphics[width=0.98\linewidth]{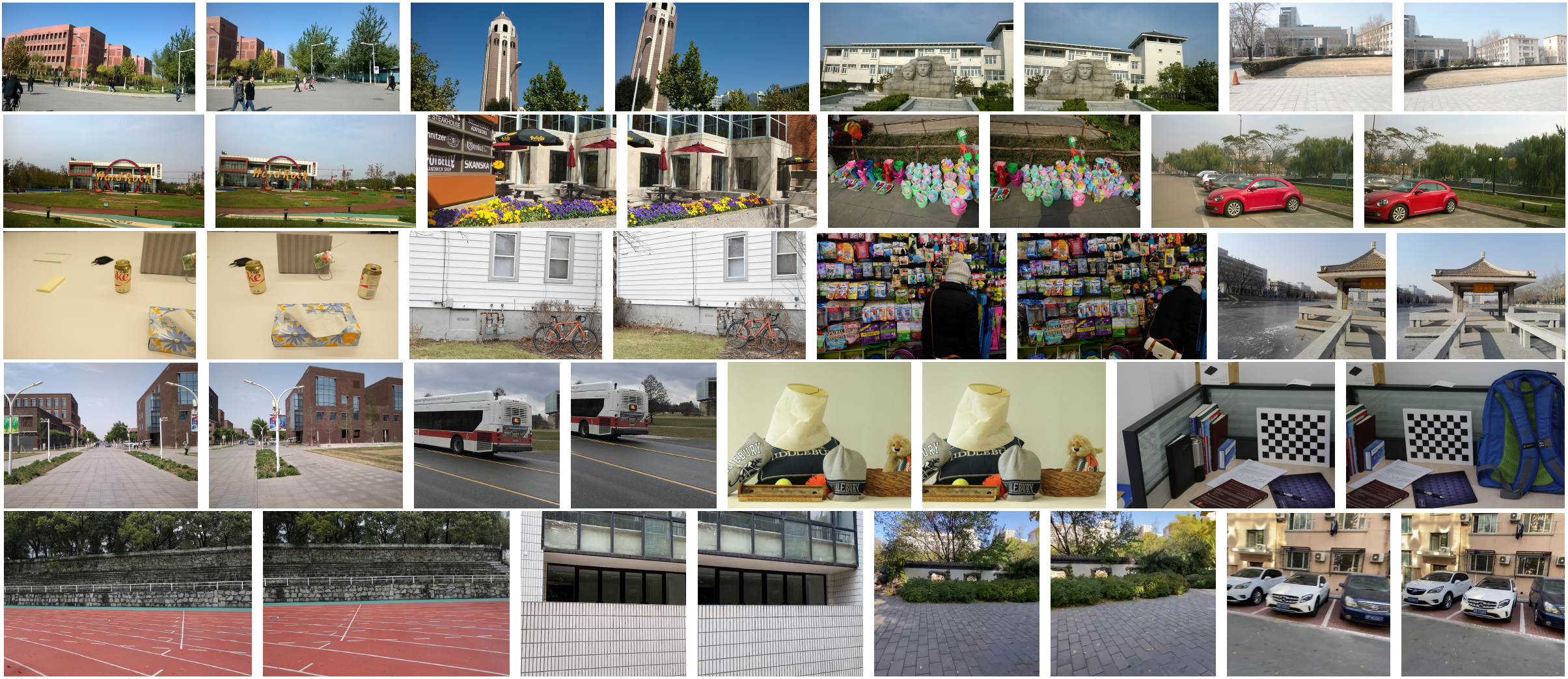}\\
	\caption{Our image dataset for experimental evaluation.}
	\label{fig: dataset}
\end{figure}

\section{Experiments}

\begin{figure*}[t]
	\centering
	\includegraphics[height=0.135\textheight]{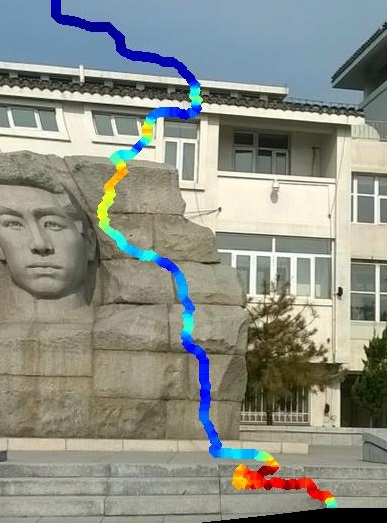}
	\includegraphics[height=0.135\textheight]{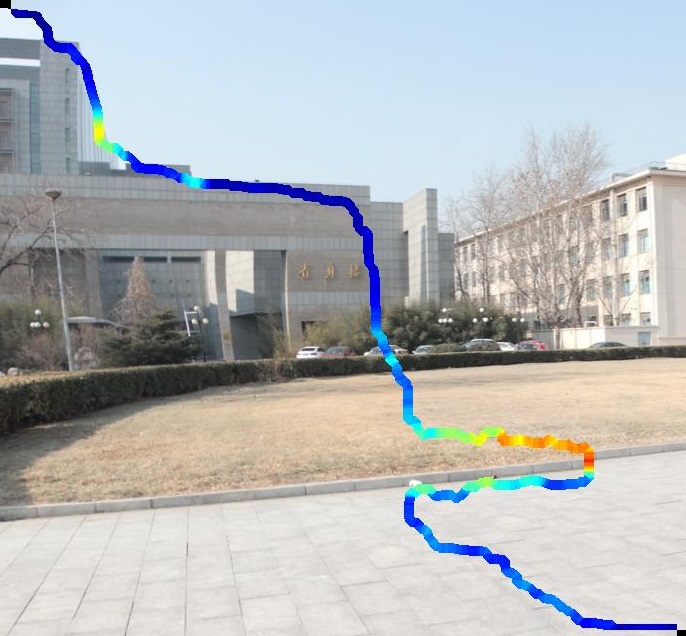}
	\includegraphics[height=0.135\textheight]{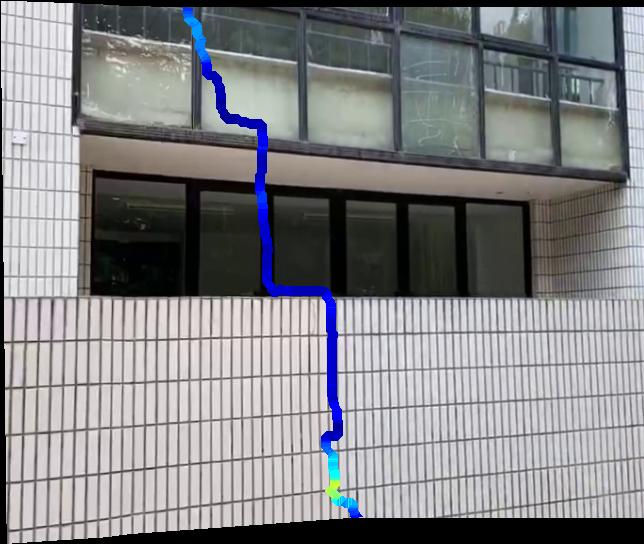}
	\includegraphics[height=0.135\textheight]{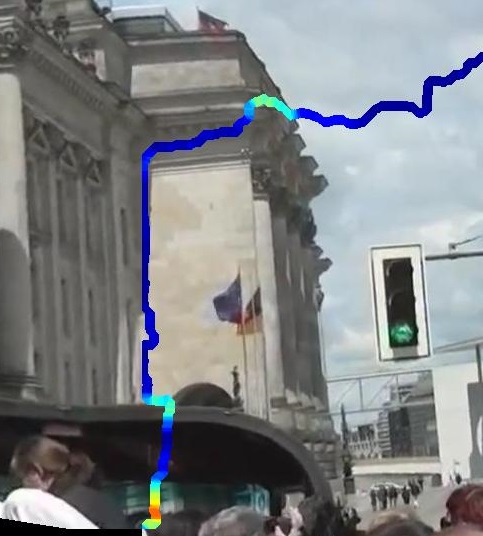}
	\includegraphics[height=0.135\textheight]{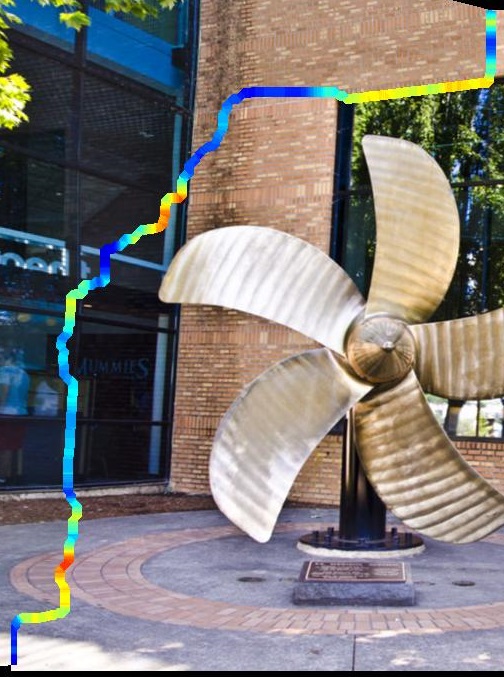}
	\includegraphics[height=0.135\textheight]{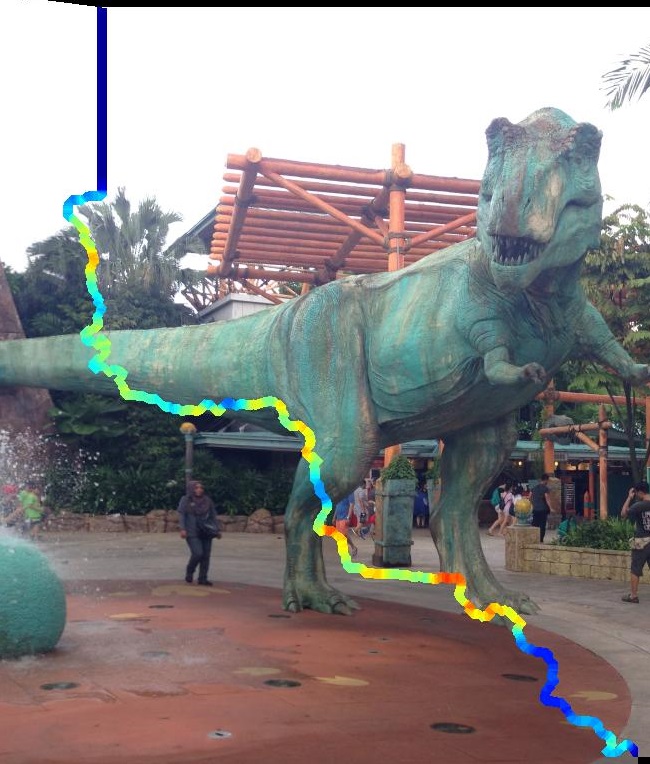}\\
	\includegraphics[height=0.135\textheight]{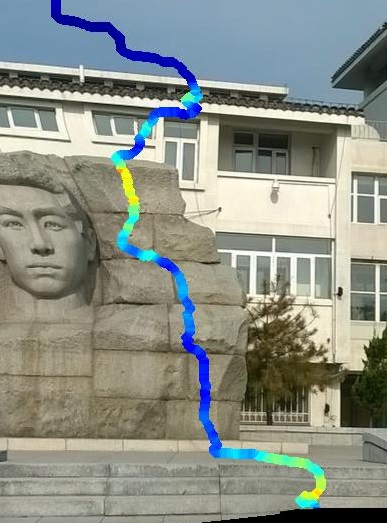}
	\includegraphics[height=0.135\textheight]{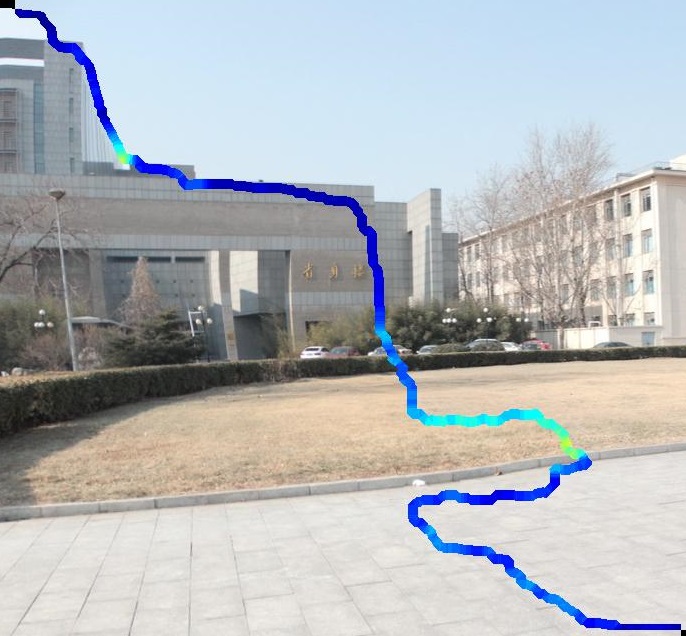}
	\includegraphics[height=0.135\textheight]{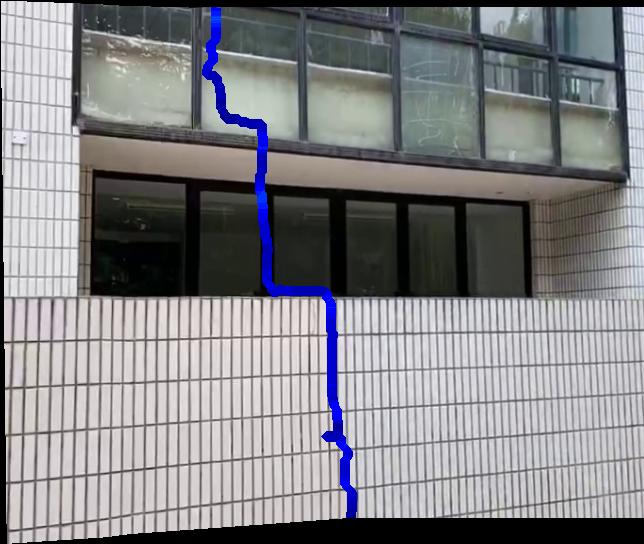}
	\includegraphics[height=0.135\textheight]{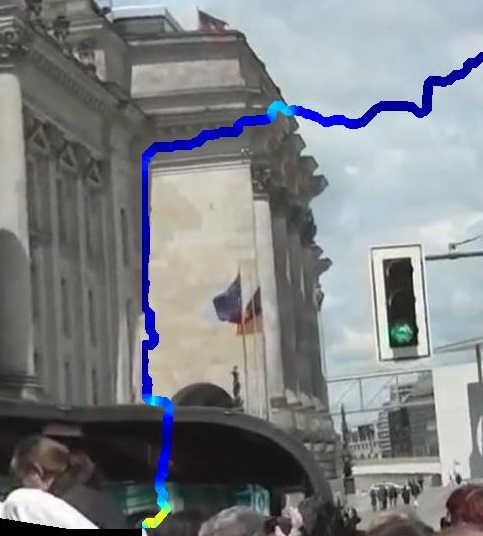}
	\includegraphics[height=0.135\textheight]{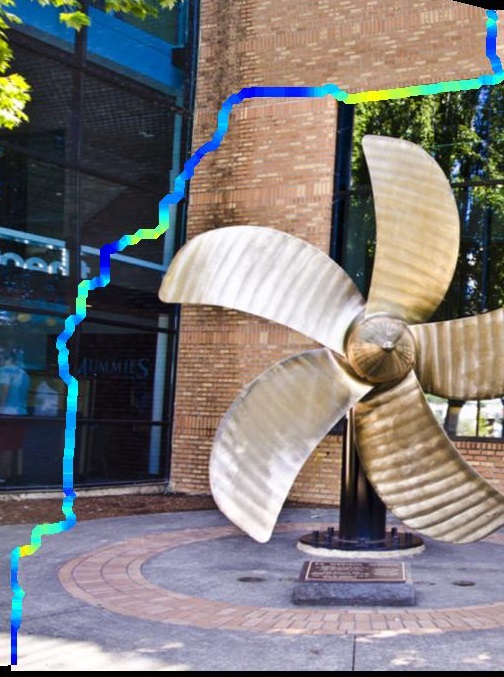}
	\includegraphics[height=0.135\textheight]{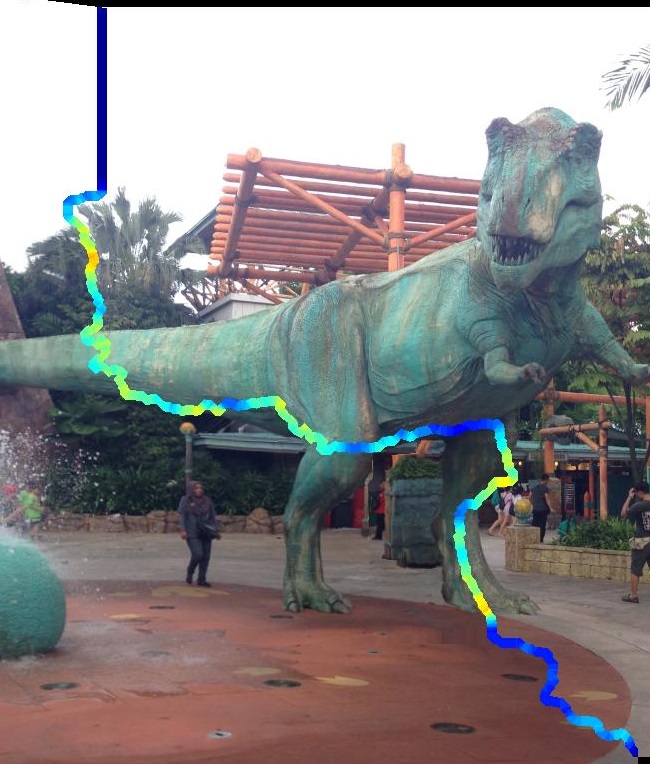}\\
	\caption{Visual seam quality comparisons w/o and w/ LPAM. Top: initial seams w/o LPAM. Bottom: final seams w/ LPAM.}
	\label{fig: examples}
\end{figure*}

In the experiments, the patch size for seam quality evaluation is set to $21\times 21$. Input images are aligned using REW \cite{li2018parallax}, a state-of-the-art warping method. To comprehensively assess the proposed method, LPAM is applied to several advanced seam-cutting methods, including basic Euclidian-based seam-cutting (\textbf{Baseline}), perception-based seam-cutting (\textbf{Perception}) \cite{li2018perception}, iterative seam-cutting (\textbf{Iterative}) \cite{liao2019Quality}, and quaternion-based seam-cutting  (\textbf{Quaternion}) \cite{li2024automatic}, the final mosaic is generated by simply compositing the aligned images according to the stitching seam. We test our method on three large parallax image stitching datasets: 35 image pairs from Parallax \cite{zhang2014parallax}, 24 image pairs from SEAGULL \cite{lin2016seagull}, and 20 image pairs from our dataset, as shown in Fig. \ref{fig: dataset}.

\begin{table*}[t]
	\centering
	\caption{Quantitative evaluation of applying LPAM to different seam-cutting methods under the same alignment results via REW \cite{li2018parallax}.}
	\label{table: comp}
	\setlength{\tabcolsep}{0.005\linewidth}{
	\small
	\begin{tabular}{lcccccccccccc}
				\toprule
				\multirow{2}[4]{*}{Dataset} & \multicolumn{4}{c}{Parallax \cite{zhang2014parallax}} & \multicolumn{4}{c}{SEAGULL \cite{lin2016seagull}} & \multicolumn{4}{c}{Our dataset} \\
				\cmidrule{2-13}          & RMSE $\downarrow$ & PSNR $\uparrow$ & SSIM $\uparrow$ & ZNCC $\downarrow$ & RMSE $\downarrow$ & PSNR $\uparrow$ & SSIM $\uparrow$ & ZNCC $\downarrow$ & RMSE $\downarrow$ & PSNR $\uparrow$ & SSIM $\uparrow$ & ZNCC $\downarrow$ \\
				\midrule
				Baseline & 0.088 & 31.01 & 0.750 & 0.119 & 0.106 & 26.20 & 0.638 & 0.163 & 0.079 & 28.50 & 0.756 & 0.114 \\
				+LPAM & \textbf{0.078} & \textbf{31.85} & \textbf{0.782} & \textbf{0.100} & \textbf{0.096} & \textbf{28.84} & \textbf{0.697} & \textbf{0.137} & \textbf{0.075} & \textbf{28.76} & \textbf{0.778} & \textbf{0.104} \\
				\midrule
				Perception \cite{li2018perception} & 0.084 & 30.03 & 0.749 & 0.102 & 0.114 & 24.43 & 0.613 & 0.159 & 0.081 & 28.11 & 0.756 & 0.106 \\
				+LPAM & \textbf{0.074} & \textbf{31.23} & \textbf{0.795} & \textbf{0.084} & \textbf{0.103} & \textbf{25.37} & \textbf{0.682} & \textbf{0.127} & \textbf{0.072} & \textbf{28.82} & \textbf{0.801} & \textbf{0.084} \\
				\midrule
				Iterative \cite{liao2019Quality} & 0.086 & 29.83 & 0.762 & 0.107 & 0.114 & 24.43 & 0.613 & 0.159 & 0.081 & 28.46 & 0.761 & 0.113 \\
				+LPAM & \textbf{0.076} & \textbf{30.87} & \textbf{0.800} & \textbf{0.087} & \textbf{0.103} & \textbf{25.37} & \textbf{0.682} & \textbf{0.127} & \textbf{0.072} & \textbf{28.91} & \textbf{0.786} & \textbf{0.101} \\
				\midrule
				Quaternion \cite{li2024automatic} & 0.089 & 28.70 & 0.746 & 0.099 & 0.116 & 25.00 & 0.613 & 0.167 & 0.073 & 28.50 & 0.778 & 0.089 \\
				+LPAM & \textbf{0.080} & \textbf{29.41} & \textbf{0.783} & \textbf{0.085} & \textbf{0.105} & \textbf{26.09} & \textbf{0.677} & \textbf{0.136} & \textbf{0.067} & \textbf{29.25} & \textbf{0.812} & \textbf{0.078} \\
				\bottomrule
	\end{tabular}}
\end{table*}
	

\begin{table*}[t]
	\centering
	\caption{Quantitative evaluation of applying LPAM to basic Euclidian-based seam-cutting (\textbf{Baseline}) under different alignment results.}
	\label{table: alignment}
	\setlength{\tabcolsep}{0.005\linewidth}{
	\small
	\begin{tabular}{lcccccccccccc}
				\toprule
				\multirow{2}[4]{*}{Dataset} & \multicolumn{4}{c}{Parallax \cite{zhang2014parallax}} & \multicolumn{4}{c}{SEAGULL \cite{lin2016seagull}} & \multicolumn{4}{c}{Our dataset} \\
				\cmidrule{2-13}          & RMSE $\downarrow$ & PSNR $\uparrow$ & SSIM $\uparrow$ & ZNCC $\downarrow$ & RMSE $\downarrow$ & PSNR $\uparrow$ & SSIM $\uparrow$ & ZNCC $\downarrow$ & RMSE $\downarrow$ & PSNR $\uparrow$ & SSIM $\uparrow$ & ZNCC $\downarrow$ \\
				\midrule
				Homo & 0.103 & 28.17 & 0.692 & 0.157 & 0.123 & 25.79 & 0.622 & 0.192 & 0.104 & 25.71 & 0.706 & 0.134 \\
				+LPAM & \textbf{0.092} & \textbf{28.78} & \textbf{0.725} & \textbf{0.133} & \textbf{0.107} & \textbf{28.50} & \textbf{0.673} & \textbf{0.165} & \textbf{0.094} & \textbf{26.22} & \textbf{0.742} & \textbf{0.114} \\
				\midrule
				ANAP \cite{lin2015adaptive} & 0.091 & 27.99 & 0.718 & 0.150 & 0.105 & 26.07 & 0.669 & 0.183 & 0.087 & 27.11 & 0.704 & 0.137 \\
				+LPAM & \textbf{0.081} & \textbf{28.64} & \textbf{0.766} & \textbf{0.123} & \textbf{0.093} & \textbf{26.88} & \textbf{0.726} & \textbf{0.146} & \textbf{0.080} & \textbf{27.73} & \textbf{0.749} & \textbf{0.112} \\
				\midrule
				GSP \cite{chen2016natural} & 0.078 & 30.14 & 0.762 & 0.105 & 0.102 & 26.34 & 0.670 & 0.161 & 0.081 & 28.36 & 0.746 & 0.117 \\
				+LPAM & \textbf{0.072} & \textbf{30.65} & \textbf{0.798} & \textbf{0.088} & \textbf{0.092} & \textbf{29.53} & \textbf{0.719} & \textbf{0.133} & \textbf{0.077} & \textbf{28.74} & \textbf{0.768} & \textbf{0.108} \\
				\midrule
                REW \cite{li2018parallax} & 0.088 & 31.01 & 0.750 & 0.119 & 0.106 & 26.20 & 0.638 & 0.163 & 0.079 & 28.50 & 0.756 & 0.114 \\
				+LPAM & \textbf{0.078} & \textbf{31.85} & \textbf{0.782} & \textbf{0.100} & \textbf{0.096} & \textbf{28.84} & \textbf{0.697} & \textbf{0.137} & \textbf{0.075} & \textbf{28.76} & \textbf{0.778} & \textbf{0.104} \\
				\midrule
				UDIS++ \cite{nie2023parallax} & 0.120 & 27.05 & 0.643 & 0.189 & 0.120 & 25.58 & 0.589 & 0.212 & 0.099 & 26.34 & 0.658 & 0.164 \\
				+LPAM & \textbf{0.100} & \textbf{28.33} & \textbf{0.705} & \textbf{0.149} & \textbf{0.109} & \textbf{26.32} & \textbf{0.651} & \textbf{0.174} & \textbf{0.088} & \textbf{27.04} & \textbf{0.715} & \textbf{0.130} \\
				\bottomrule
	\end{tabular}}
\end{table*}

\subsection{Quantitative evaluation}

We introduce four classical metrics, RMSE, PSNR, SSIM \cite{wang2004image}, and ZNCC \cite{lin2016seagull} to evaluate the quality of the initial seam and final seam after adding the LPAM. For the stitching seam $S$, the quality errors are defined as
\begin{gather}
	\mathrm{RMSE}(S)=\frac{1}{N}\sum_{i=1}^N\mathrm{RMSE}(P_0(p_i), P_1(p_i))\label{eq:quality1}\\
	\mathrm{PSNR}(S)=\frac{1}{N}\sum_{i=1}^N\mathrm{PSNR}(P_0(p_i), P_1(p_i))\label{eq:quality2}\\
	\mathrm{SSIM}(S)=\frac{1}{N}\sum_{i=1}^N\mathrm{SSIM}(P_0(p_i), P_1(p_i))\label{eq:quality3}\\
	\mathrm{ZNCC}(S)=\frac{1}{N}\sum_{i=1}^N\frac{1-\mathrm{ZNCC}(P_0(p_i), P_1(p_i))}{2}\label{eq:quality4}
\end{gather}
where $p_i$ is the pixel on the seam, $N$ is the number of the pixels on the seam, $P_0$, $P_1$ is the patch centered at $p_i$ in $I_0$, $I_1$ respectively. 

\subsubsection{Evaluation across different seam-cutting methods}
Table \ref{table: comp} presents the quantitative evaluation results of adding LPAM to different seam-cutting methods under the same alignment results via REW \cite{li2018parallax}. Evidently, LPAM can substantially improve the seam quality across different seam-cutting methods on all large parallax stitching datasets. We also show part of the seam comparison results in Fig. \ref{fig: examples}, where the seams are shown using colormaps based on the quality error defined by Eq. (\ref{eq:ssim}). Through the local alignment of misaligned patches, our method effectively improves the quality of seam-cutting and the final visual stitching results.


\subsubsection{Evaluation across different alignment methods}

As shown in Table \ref{table: alignment}, we also evaluate the LPAM when applying it to a single seam-cutting method under the different representative alignment methods, including global homography (Homo), ANAP \cite{lin2015adaptive}, GSP \cite{chen2016natural}, REW \cite{li2018parallax} and the state-of-the-art learning-based method UDIS++ \cite{nie2023parallax}. LPAM still achieves stable improvement of the seam quality across different alignment results.

\begin{table*}[t]
	\centering
	\caption{Ablation study of applying LPAM to basic Euclidian-based seam-cutting. Alignment results are obtained via REW \cite{li2018parallax}.}
	\label{table: ablation}
	\setlength{\tabcolsep}{0.005\linewidth}{
	\small
	\begin{tabular}{ccccccccccccc}
				\toprule
				\multirow{2}[4]{*}{Dataset} & \multicolumn{4}{c}{Parallax \cite{zhang2014parallax}} & \multicolumn{4}{c}{SEAGULL \cite{lin2016seagull}} & \multicolumn{4}{c}{Our dataset} \\
				\cmidrule{2-13} & RMSE $\downarrow$ & PSNR $\uparrow$ & SSIM $\uparrow$ & ZNCC $\downarrow$ & RMSE $\downarrow$ & PSNR $\uparrow$ & SSIM $\uparrow$ & ZNCC $\downarrow$ & RMSE $\downarrow$ & PSNR $\uparrow$ & SSIM $\uparrow$ & ZNCC $\downarrow$ \\
				\midrule
                $\beta=\emptyset$ & 0.069 & 31.38 & 0.819 & 0.082 & 0.084 & 28.24 & 0.768 & 0.101 & 0.069 & 29.33& 0.815 & 0.086 \\
				$\beta=1$ & 0.080 & 30.51 & 0.770 & 0.103 & 0.099  & 28.75 & 0.677 & 0.143 & 0.076 & 28.73 & 0.774 & 0.103 \\
                $\beta=2$ & 0.080 & 30.49 & 0.771 & 0.103 & 0.098 & 28.88 & 0.680 & 0.142 & 0.076 & 28.73 & 0.774 & 0.105 \\
                $\beta=4$ & 0.080 & 30.73 & 0.775 & 0.102 & 0.098 & 28.23 & 0.683 & 0.142 & 0.076 & 28.72 & 0.771 & 0.106 \\
                \rowcolor{gray!40}
                $\beta=8$ & 0.078 & 31.85 & 0.782 & 0.100 & 0.096 & 28.84 & 0.697 & 0.137 & 0.075 & 28.76 & 0.778 & 0.104 \\
                $\beta=16$ & 0.078 & 30.92 & 0.784 & 0.099 & 0.095 & 29.30 & 0.706 & 0.132 & 0.075 & 28.84 & 0.782 & 0.103 \\
                \midrule
                $7\times 7$ & 0.084	& 29.90 & 0.762	& 0.110 & 0.099	& 28.05 & 0.679 & 0.145 & 0.076	& 28.66 & 0.774 & 0.107\\
                $11\times 11$ & 0.081 &	30.13 & 0.771 & 0.104 & 0.098 & 28.16 & 0.683 & 0.142 & 0.075 & 28.74 & 0.776 & 0.106\\
                $31\times 31$ & 0.078 & 31.69 & 0.783 & 0.099 & 0.098 & 27.52 & 0.692 & 0.137 & 0.076 & 28.78 & 0.784 & 0.102\\
                $41\times 41$ & 0.078 & 31.67 & 0.777 & 0.099 & 0.100 & 27.39 & 0.684 & 0.140 & 0.074 & 28.93 & 0.784 & 0.103\\
                $51\times 51$ & 0.081 & 31.80 & 0.775 & 0.100 & 0.099 & 28.05 & 0.684 & 0.144 & 0.075 & 28.91 & 0.782 & 0.103\\
				\bottomrule
	\end{tabular}}
\end{table*}

\subsection{Visual results comparison}

\begin{figure*}
	\centering
	\subfloat[\centering Input images]{
		\includegraphics[height=0.1087\textheight]{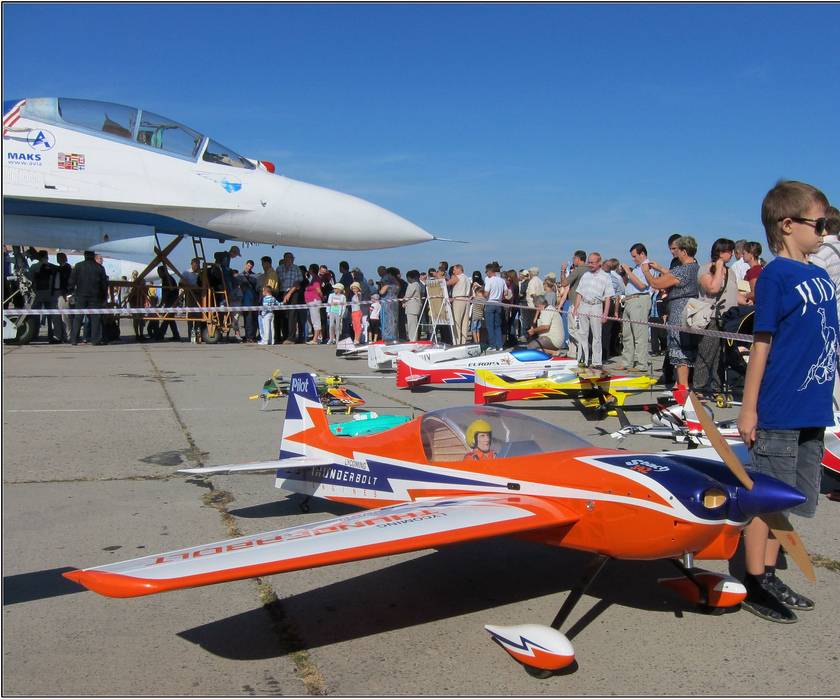}
		\includegraphics[height=0.1087\textheight]{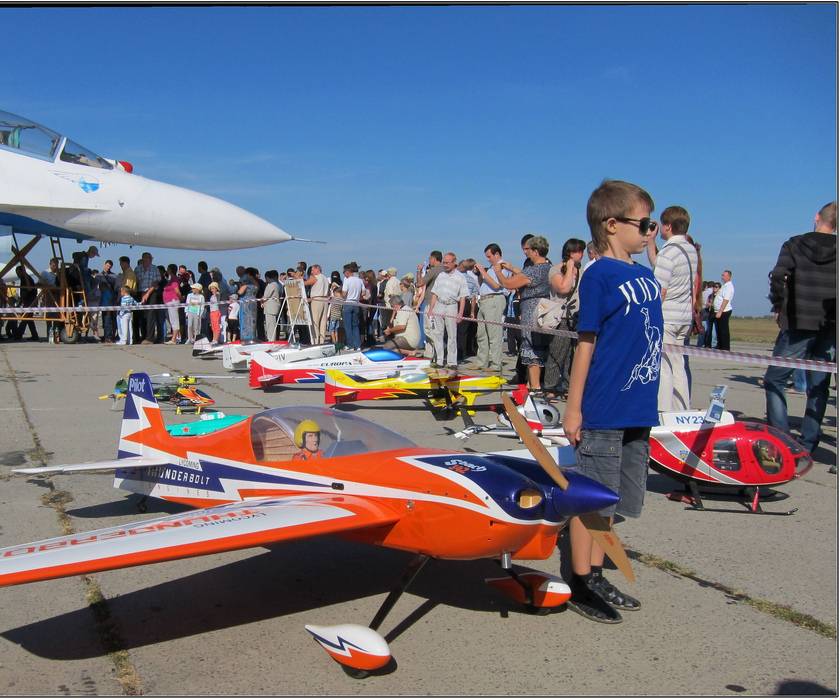}}
	\subfloat[\centering Microsoft ICE]{
		\includegraphics[height=0.1087\textheight]{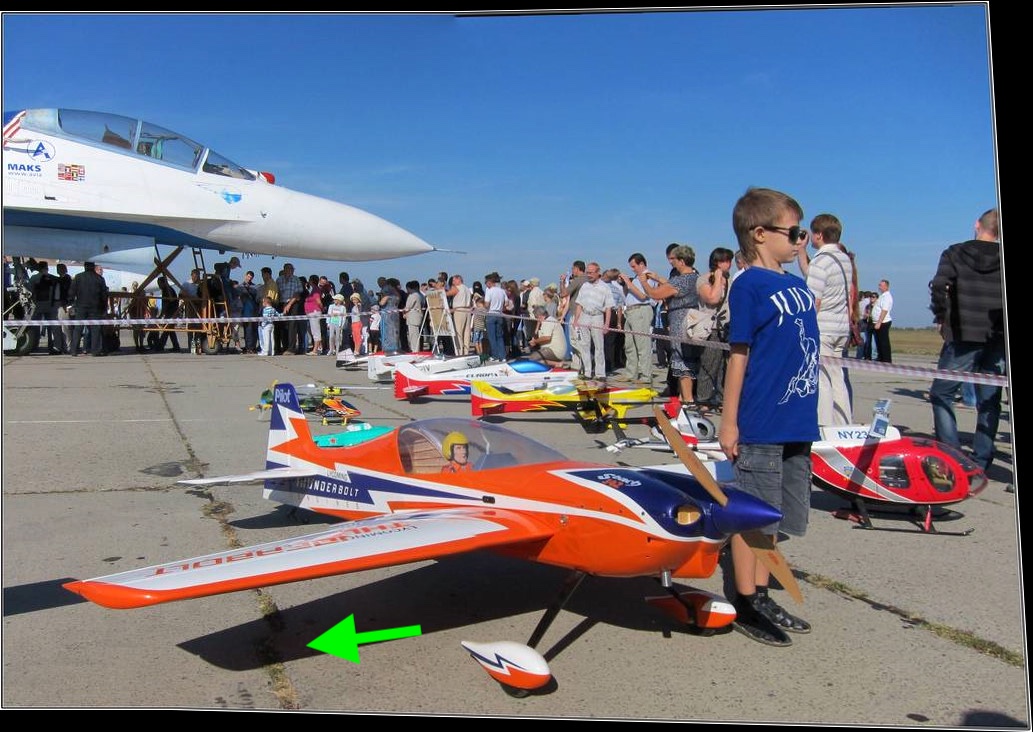}}
	\subfloat[\centering Parallax \cite{zhang2014parallax}]{
		\includegraphics[height=0.1087\textheight]{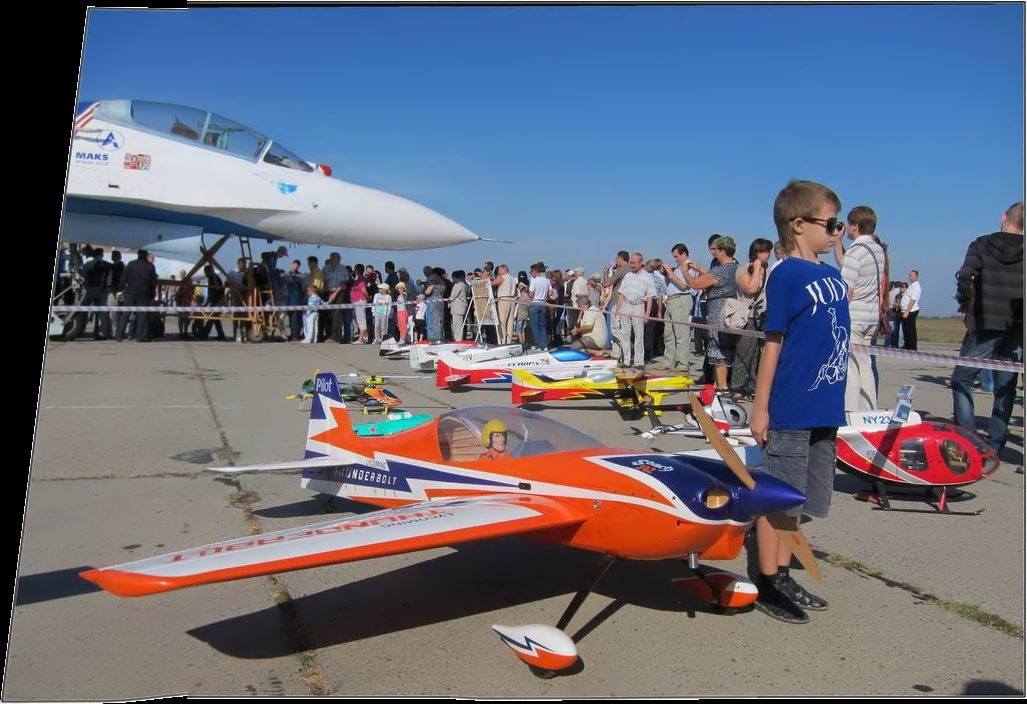}}
	\subfloat[\centering SEAGULL \cite{lin2016seagull}]{
		\includegraphics[height=0.1087\textheight]{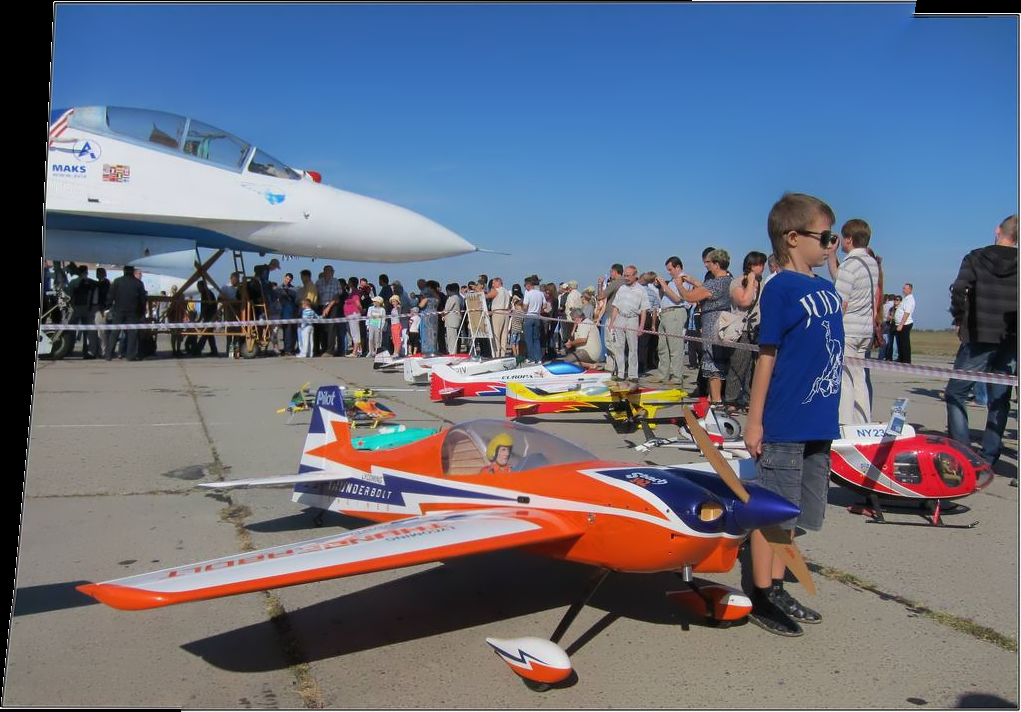}}\\
	\subfloat[\centering Baseline]{
		\includegraphics[height=0.13\textheight]{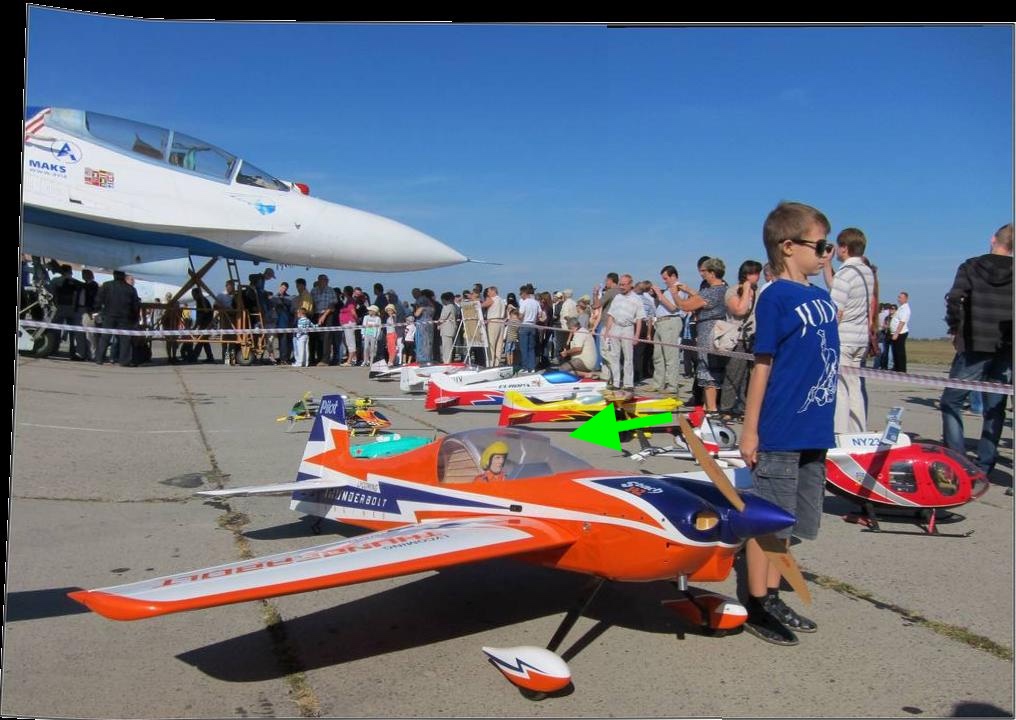}
		\includegraphics[height=0.13\textheight]{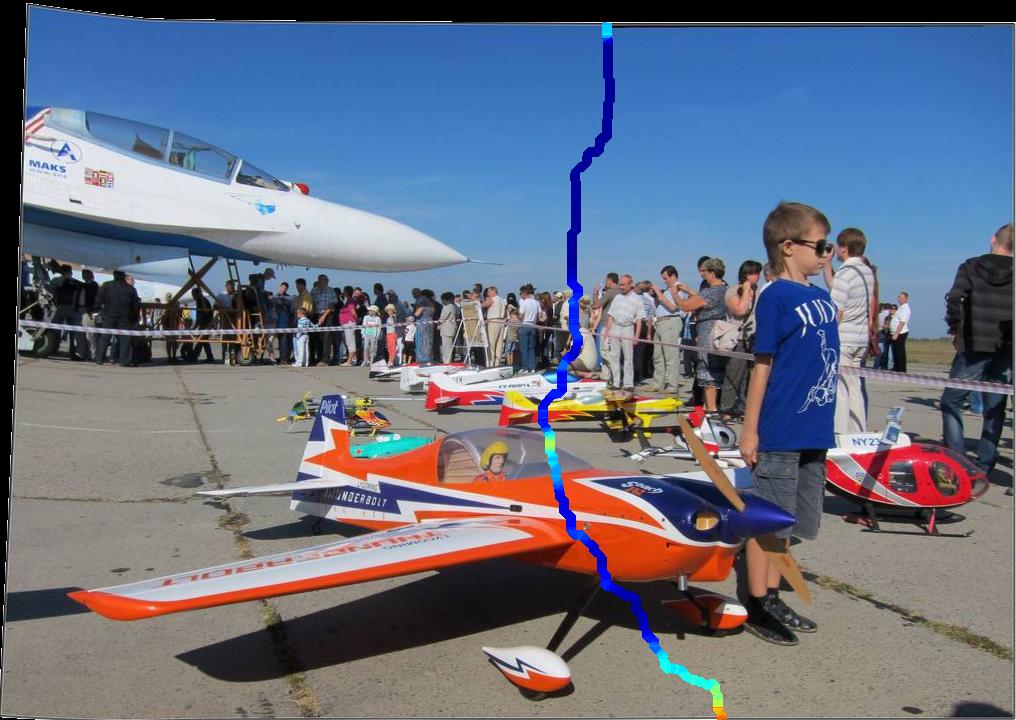}}
	\subfloat[\centering Baseline+LPAM]{
		\includegraphics[height=0.13\textheight]{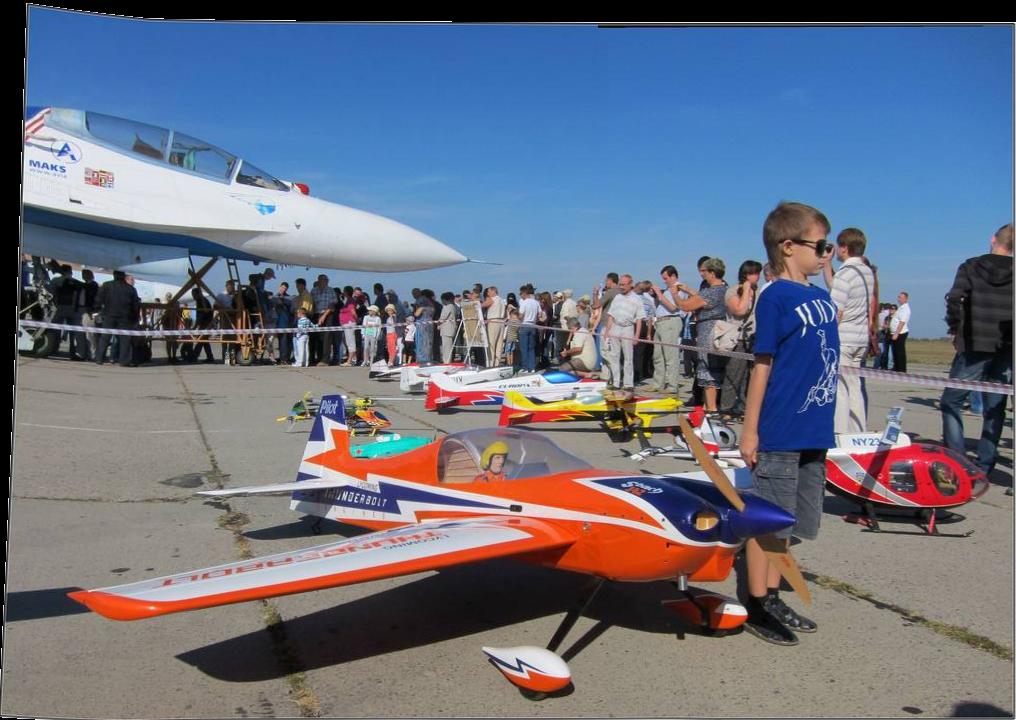}
		\includegraphics[height=0.13\textheight]{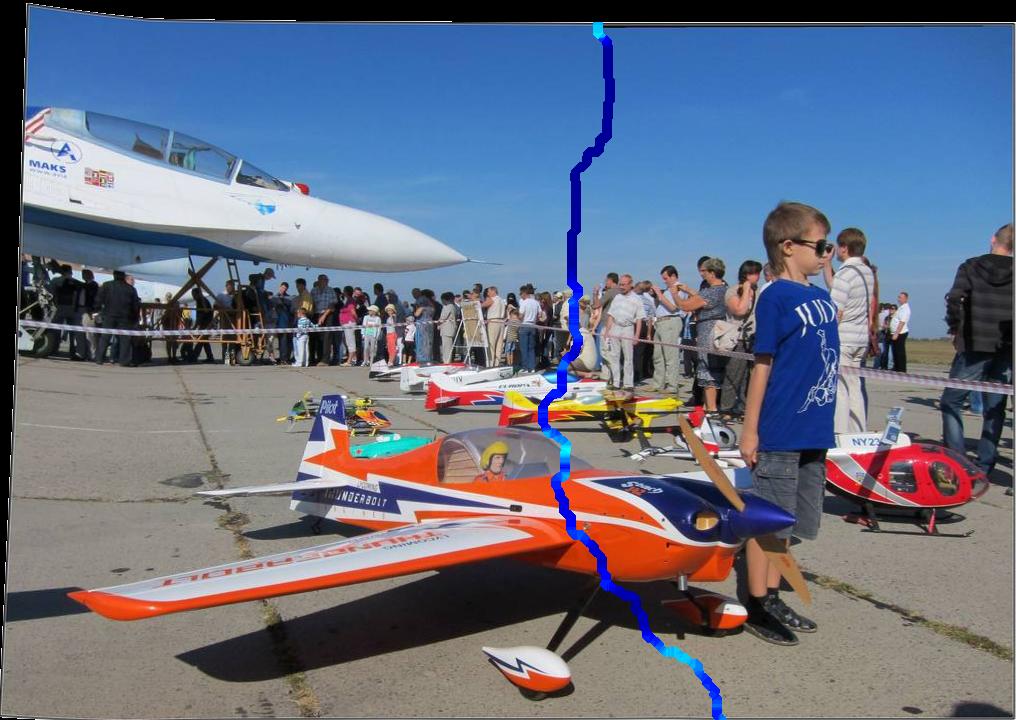}}\\
	\subfloat[\centering Perception \cite{li2018perception}]{
		\includegraphics[height=0.13\textheight]{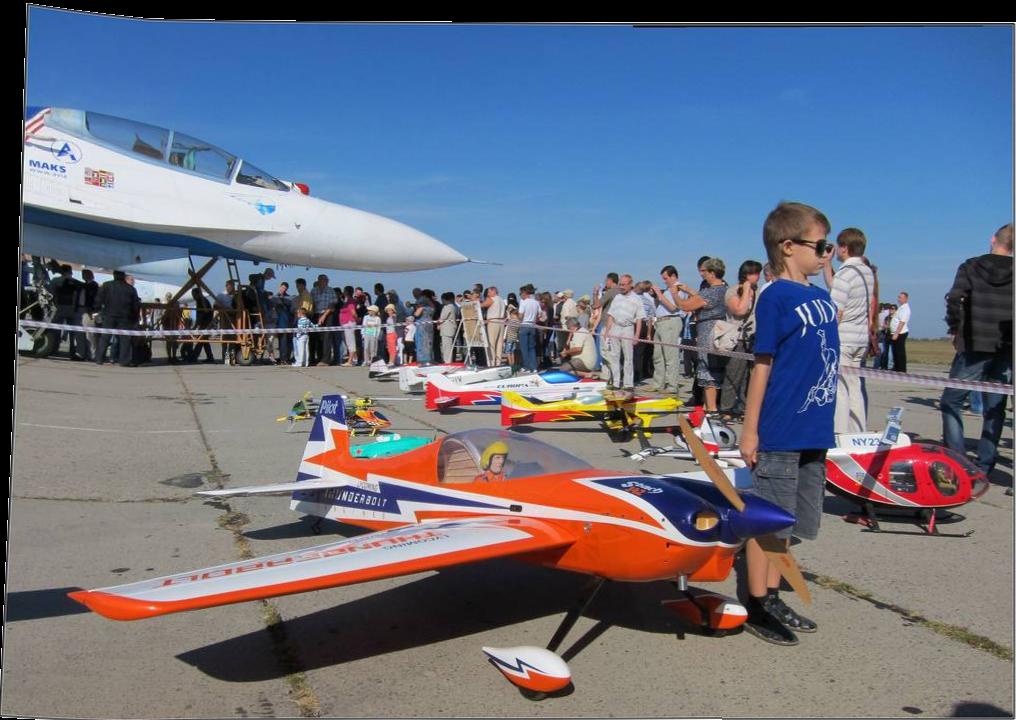}
		\includegraphics[height=0.13\textheight]{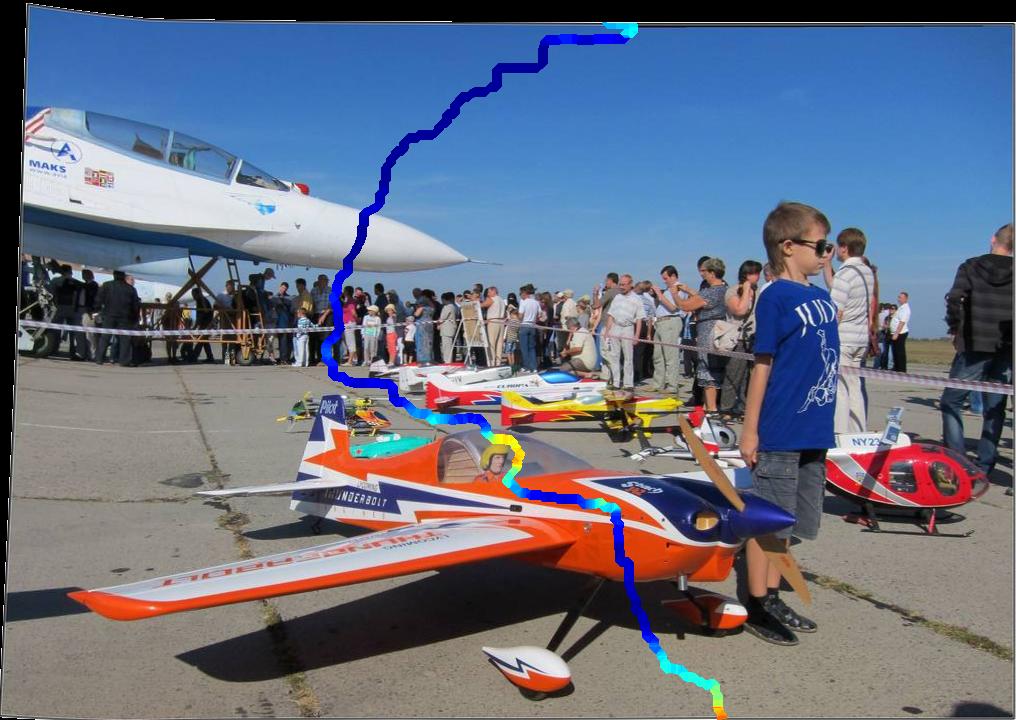}}
	\subfloat[\centering Perception+LPAM]{
		\includegraphics[height=0.13\textheight]{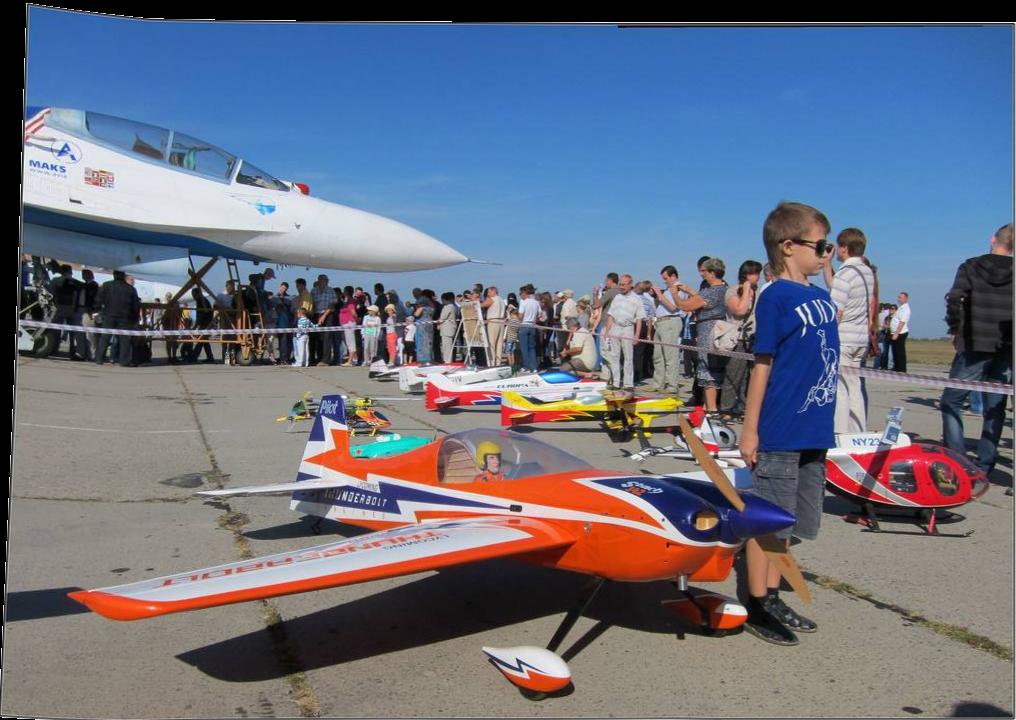}
		\includegraphics[height=0.13\textheight]{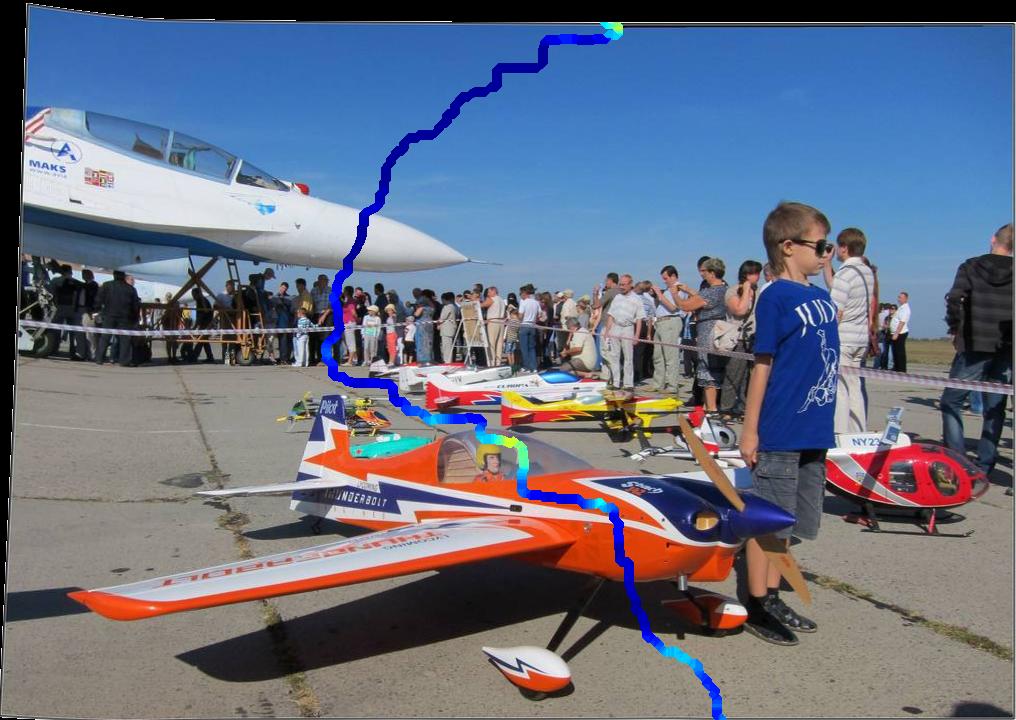}}\\
	\subfloat[\centering Iterative \cite{liao2019Quality}]{
		\includegraphics[height=0.13\textheight]{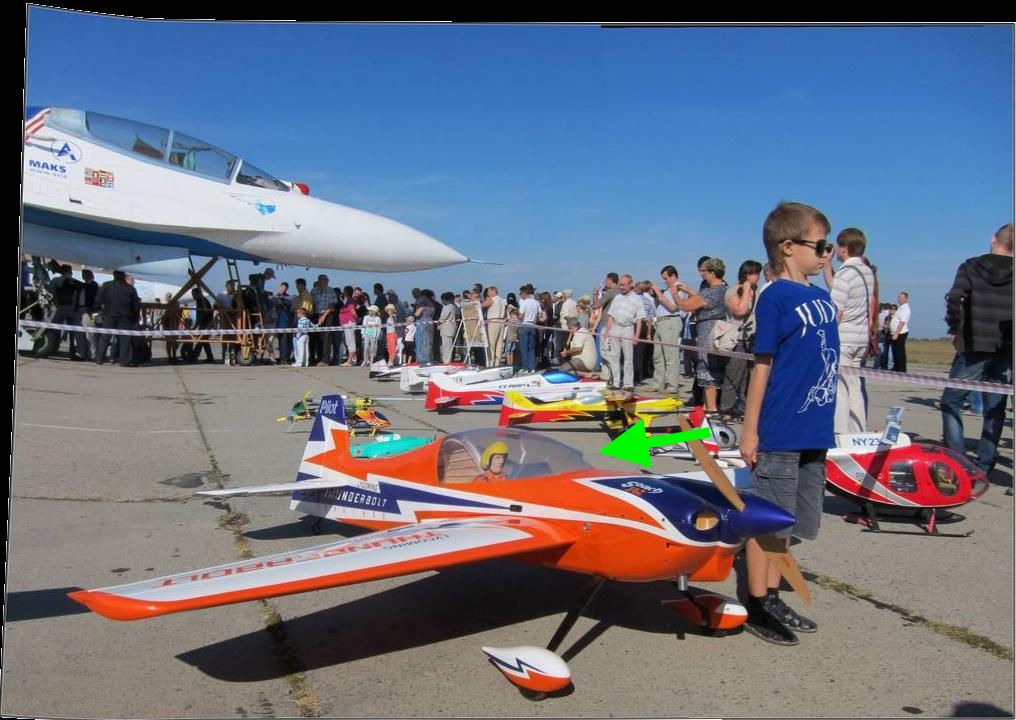}
		\includegraphics[height=0.13\textheight]{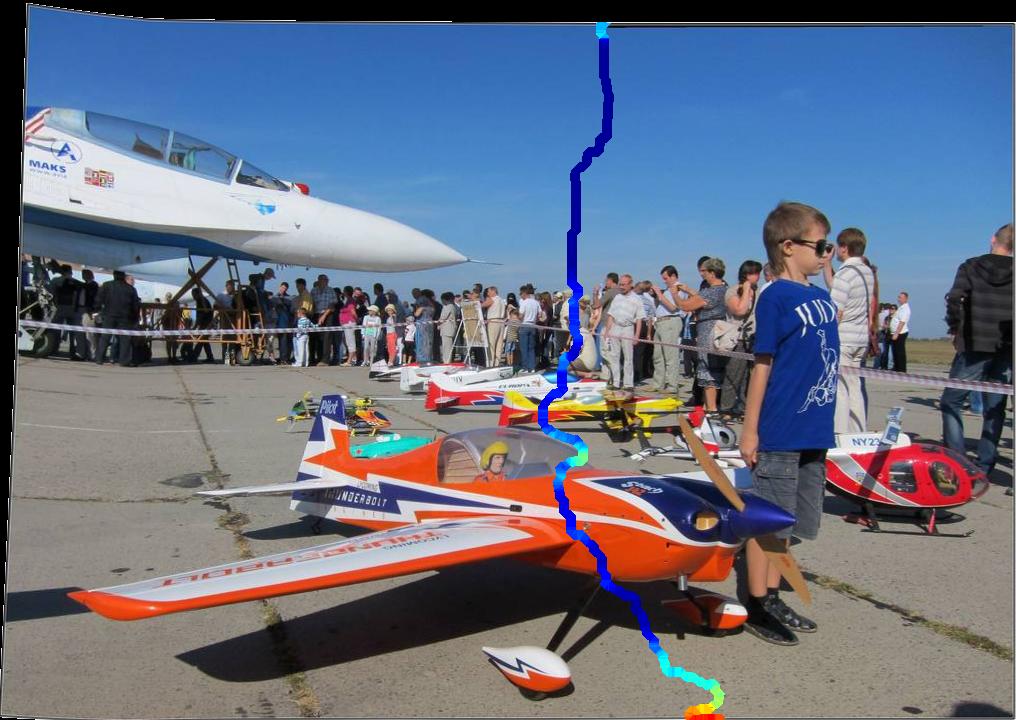}}
	\subfloat[\centering Iterative+LPAM]{
		\includegraphics[height=0.13\textheight]{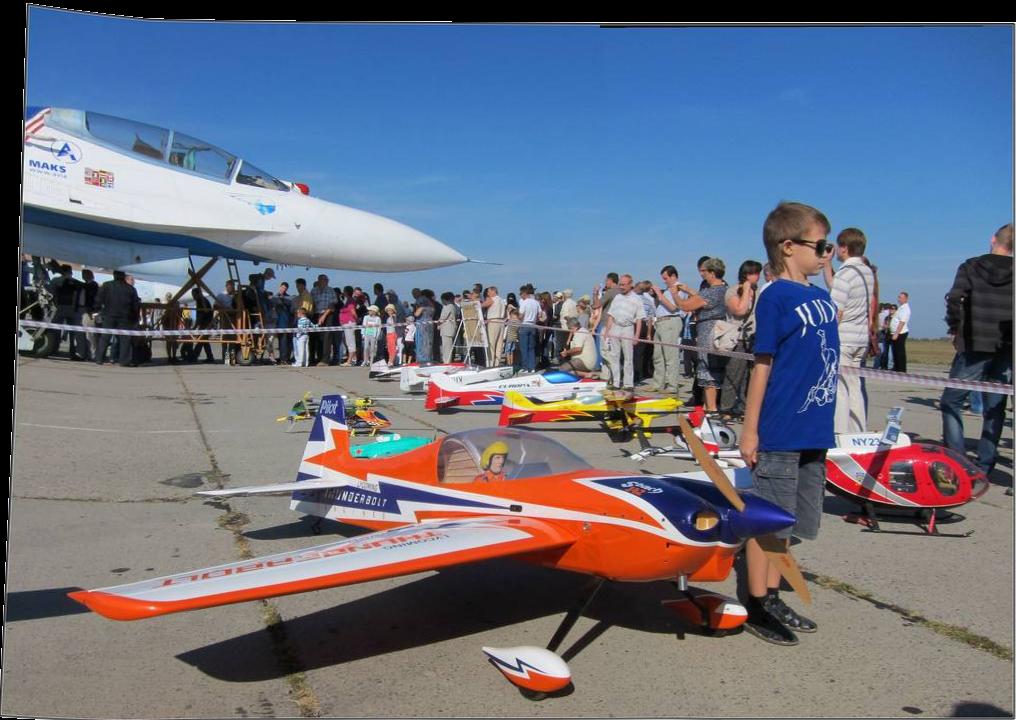}
		\includegraphics[height=0.13\textheight]{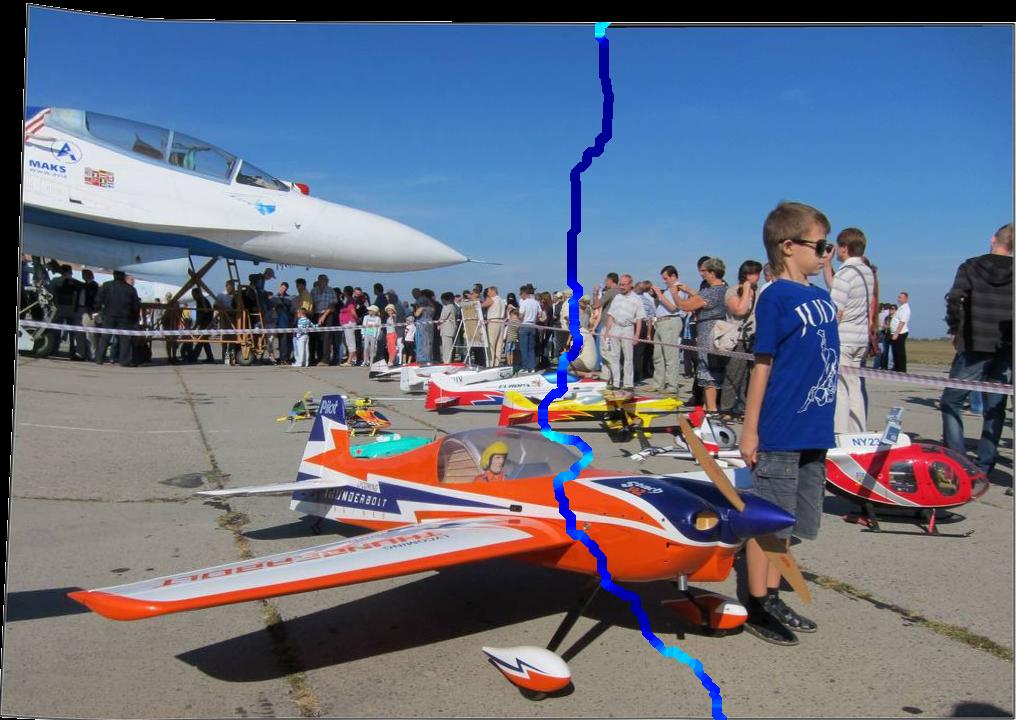}}\\
	\subfloat[\centering Quaternion \cite{li2024automatic}]{
		\includegraphics[height=0.13\textheight]{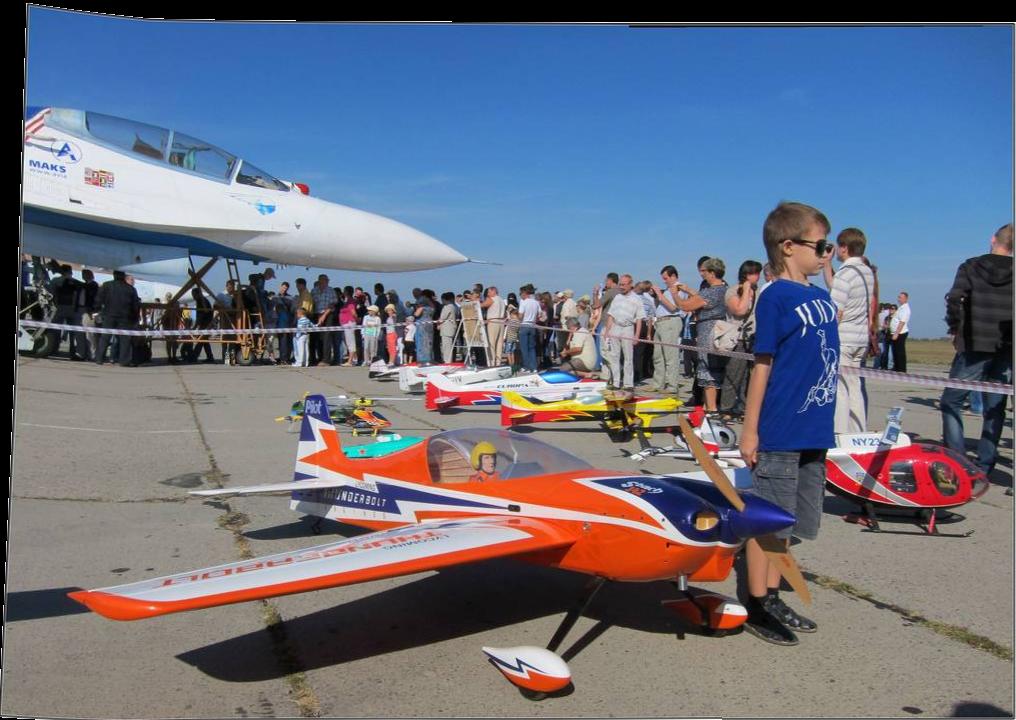}
		\includegraphics[height=0.13\textheight]{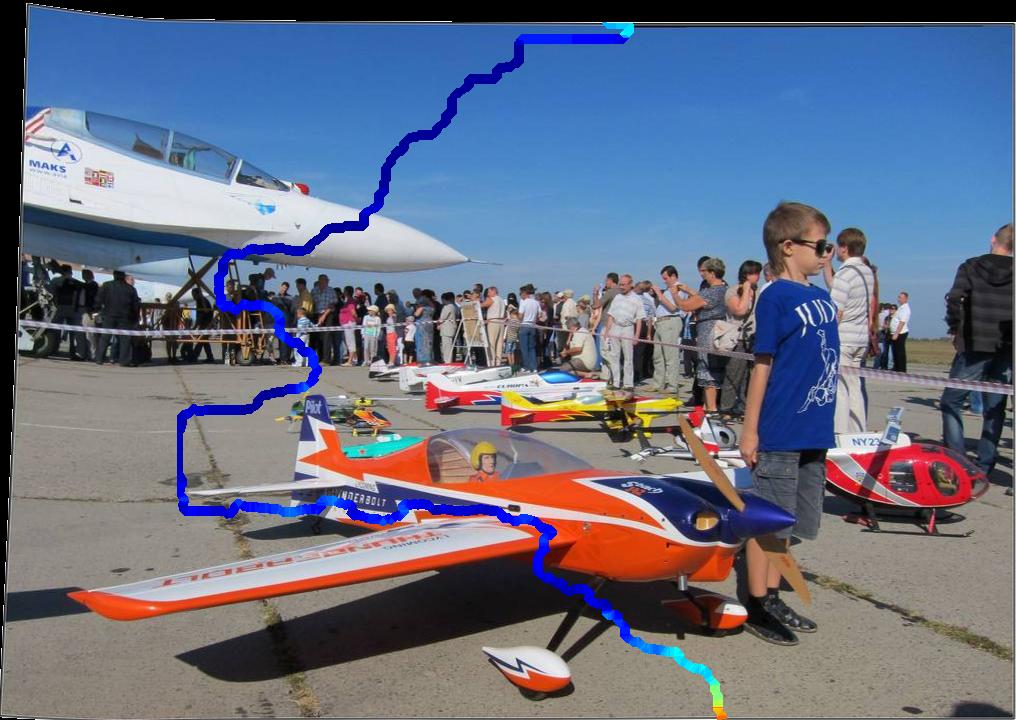}}
	\subfloat[\centering Quaternion+LPAM]{
		\includegraphics[height=0.13\textheight]{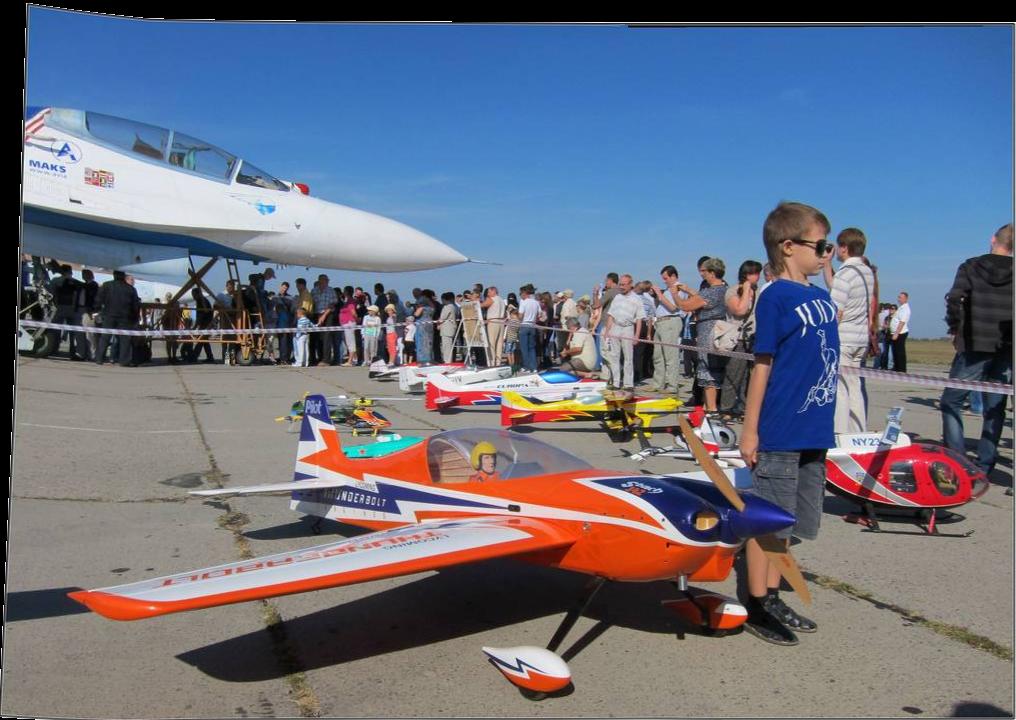}
		\includegraphics[height=0.13\textheight]{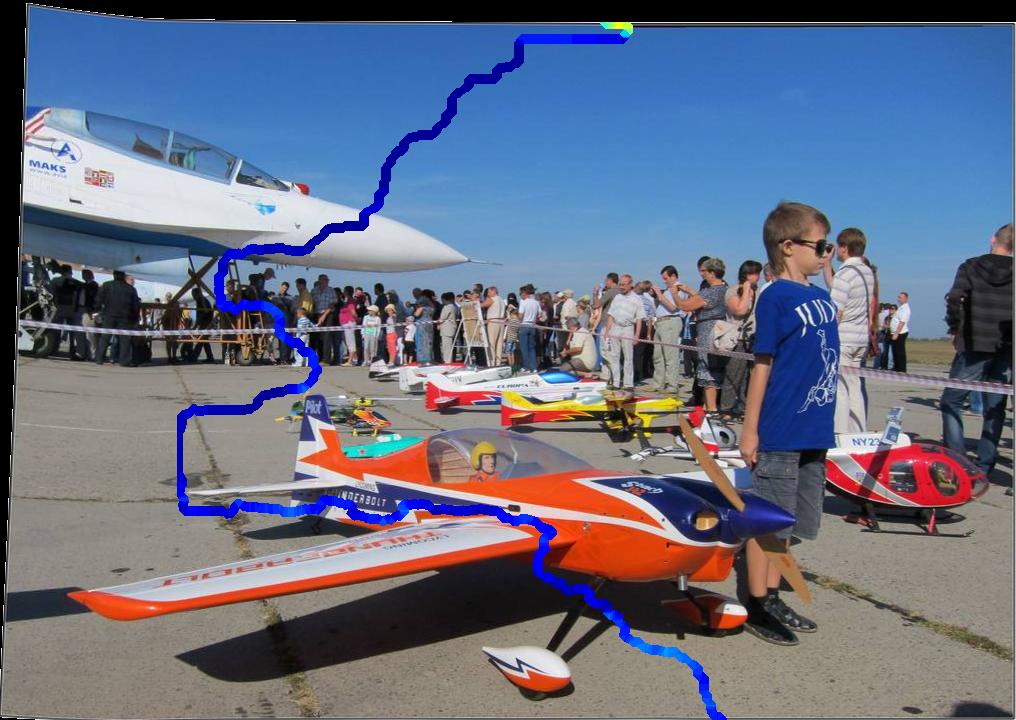}}\\
	\caption{Comparisons of image stitching results via different seam-cutting methods. (Best view in color and
zoom in)}
	\label{fig: comp}
\end{figure*}

We comprehensively compare the visual stitching results of our method with those of methods integrating the seam-cutting technique. These include Microsoft ICE\footnote{\url{https://www.microsoft.com/en-us/research/project/image-composite-editor}}, \cite{li2018perception, liao2019Quality, li2024automatic}, Parallax \cite{zhang2014parallax}, and SEAGULL \cite{lin2016seagull}. 
Fig. \ref{fig: comp} shows a particularly challenging example sourced from \cite{zhang2014parallax}. The input images and the result of Parallax are retrieved from the website\footnote{\url{http://web.cecs.pdx.edu/~fliu/project/stitch/}}. 
Notably, Microsoft ICE, the baseline method, and Iterative seam-cutting \cite{liao2019Quality} fail to produce satisfactory results (obvious artifacts are marked by green arrows). 
The simple integration of our LPAM improves the stitching quality and achieves visual results comparable to those of Parallax and SEAGULL.
It should be noted that Parallax and SEAGULL need to estimate multiple seam-cutting results from multiple alignment candidates and select the optimal one. In contrast, our LPAM provides a more straightforward and effective solution for improving stitching quality. More visual comparisons are provided in the supplementary.

\begin{figure}[t]
	\centering
	\subfloat[\centering Stitching seam and result w/o LPAM]{
		\includegraphics[height=0.14\textheight]{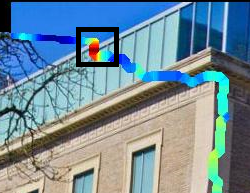}
		\includegraphics[height=0.14\textheight]{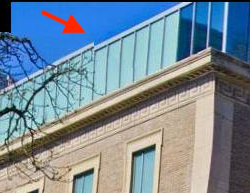}}\\
	\subfloat[\centering Stitching seam and result w/ LPAM]{
		\includegraphics[height=0.14\textheight]{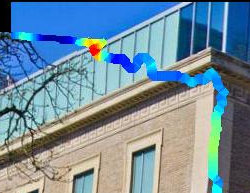}
		\includegraphics[height=0.14\textheight]{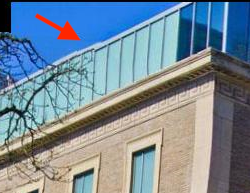}}\\
	\caption{Failure examples of our method (Best view in color and zoom in).}
	\label{fig: failure}
\end{figure}

\subsection{Ablation study}			

We conduct an ablation study to assess the performance of our LPAM under different parameter settings, as shown in Table \ref{table: ablation}. The first row denotes that we ablate the sigmoid function in Eq. (\ref{eq:sigmoid}) and use the original sift flow to align the patches. The 2nd to 6th rows represent that the $\beta$ in Eq. (\ref{eq:siftflow_1}) is set to 1,2,4,8,16. The last five rows represent different patch sizes for seam quality evaluation in Eq. (\ref{eq:ssim}) (the patch size in Eqs. (\ref{eq:quality1}-\ref{eq:quality4}) is fixed at 21$\times$21 for fair comparison), including the ``gray'' row is performed using $21\times 21$ patch size. Comparison results confirm that our LPAM can achieve stable improvement under diverse parameter settings of $\beta$. The first row shows that ablating the sigmoid function improves the seam quality according to the seam quality metrics. However, it may introduce artifacts outside the separated rectangular patches, as shown in Fig. \ref{fig:siftflow}. The study also shows that our LPAM is relatively stable under different patch sizes. Small patch size may limit smooth merging, while a large one challenges alignment and costs more (we discuss this in the supplementary). Considering these, we set the patch size to $21\times21$ to balance the smooth merging and computational efficiency.

\subsection{Computational efficiency}

We evaluate the computational efficiency of LPAM when integrated with other seam-cutting methods. Experiments are conducted using a 3.6GHz Intel Core i7 CPU with 16GB of memory. We record the elapsed time for each seam-cutting method\footnote{Time recorded is solely for seam-cutting, excluding initial image alignment.} and for LPAM individually. The results are presented in Table \ref{table: time}. In LPAM, constructing dense correspondence dominates the elapsed time. This process is influenced by the number and resolution of extracted patches, which vary significantly across image pairs and datasets. Notably, when applied to various seam-cutting methods, LPAM's processing time remains relatively stable, ranging from 6 to 10 seconds\footnote{Currently, LPAM is implemented without integration with C++ for further acceleration.}. The computational cost of each process in our LPAM is discussed in the supplementary.

\begin{table}[t]
	\centering
	\caption{Elapsed time of applying LPAM to different seam-cutting methods on the three image datasets.}
	\setlength{\tabcolsep}{0.005\linewidth}{
    \small
		\begin{tabular}{lccc}
			\toprule
			Dataset & Parallax \cite{zhang2014parallax} & SEAGULL \cite{lin2016seagull} & Our dataset \\
			\midrule
			Baseline & 7.54  & 4.11  & 5.26 \\
			+LPAM & 7.77  & 7.34  & 6.89 \\
			\midrule
			Perception \cite{li2018perception} & 9.21  & 3.61  & 6.54 \\
			+LPAM & 8.96  & 10.12 & 7.90 \\
			\midrule
			Iterative \cite{liao2019Quality} & 112.04 & 201.06 & 153.07 \\
			+LPAM & 8.85  & 9.15  & 7.47 \\
			\midrule
			Quaternion \cite{li2024automatic} & 8.95  & 3.51  & 9.59 \\
			+LPAM & 9.25  & 10.17 & 9.57 \\
			\bottomrule
	\end{tabular}}
	\label{table: time}%
\end{table}

\subsection{Discussion and limitation}


In Eqs. (\ref{eq:siftflow_1},\ref{eq:sigmoid}), we modify the vector flow estimated by SIFT flow to ensure the aligned patch is smoothly consistent with pre-aligned images. However, when significant structural misalignment exists within small patches, this modification is insufficient to ensure smoothness, as shown in Fig. \ref{fig: failure}(c,d). The extracted patch (marked by the black box) is too small for effective smoothness modification. Consequently, our LPAM struggles to attain a harmonious balance between accurate patch alignment and smooth consistency. Adaptively enlarging the patch size could yield smoother consistency. We leave this as future work.

\section{Conclusion}

In this paper, we propose a simple LPAM that can effectively and efficiently enhance the stitching performance of seam-cutting methods for large parallax image stitching. Our method is founded on a crucial discovery: the seam-cutting method is a downstream process that depends on successful image alignment. We introduce an alignment-compensation paradigm to dissociate seam quality from initial image alignment. Experimental results of applying our LPAM to the state-of-the-art seam-cutting methods with diverse alignment results on various stitching datasets demonstrate the superiority and robustness of our proposed method.  

\section*{Acknowledgments}

This work is partially supported by the Natural Science Foundation of Henan Province, China under Grant 222300420140.

{
    \small
    \bibliographystyle{ieeenat_fullname}
    \bibliography{bib_submit}
}

\end{document}